# Text Complexity Classification Based on Linguistic Information: Application to Intelligent Tutoring of ESL

M. Zakaria Kurdi

University of Lynchburg, USA

* Corresponding author: kurdi_m@lynchburg.edu

**Abstract**
The goal of this work is to build a classifier that can identify text complexity within the context of teaching reading to English as a Second Language (ESL) learners. To present language learners with texts suitable to their level of English, a set of features that can describe the phonological, morphological, lexical, syntactic, discursive, and psychological complexity of a text were identified and used for classifying texts. Using a corpus of 6171 texts, which had already been classified into three different levels of difficulty by ESL experts, different experiments were conducted with five Machine Learning algorithms. The results showed that the adopted linguistic features provide a good overall classification performance (F-Score = 0.97). A scalability evaluation was conducted to test if such a classifier could be used within real applications, where it can be plugged into a search engine or a web-scraping module. In this evaluation, the texts in the test set are not only different from those from the training set, but also of different types (ESL texts vs. children reading texts). Although the overall performance of the classifier decreased significantly (F-Score = 0.65), the confusion matrix shows that most of the classification errors are between the classes two and three (the middle-level classes). Likewise, the confusion matrix also shows that the system has a robust performance in categorizing texts of class one and class four. This behavior can be explained by the difference in classification criteria between the two corpora. Hence, the observed results confirm the usability of such a classifier within a real-world application.

# I INTRODUCTION

With the recent progress in AI, building an adaptive educational recommender system to teach a foreign language became more popular. Among others, such a recommender system should be able to retrieve textual content from the web that suits the learner's current linguistic level and propose it as a reading material. Therefore, NLP and ESL specialists became interested in automatically measuring the content complexity of a text, typically using a few features about syntax or lexicon. Most of these works were at the stage of elementary exploration and no scalability experiments were conducted. Unlike the previous works, this paper assumes that the linguistic complexity of a text is a function of features from all the linguistic levels: phonology, morphology, lexicon, syntax, and discourse in addition to psycholinguistic features such as the age of acquisition of a word or its imageability. Hence, this paper provides a more comprehensive comparison of a wide number of features that cover the linguistic field. The considered features are evaluated within their linguistic area or across the linguistic areas. This in-depth evaluation of the features makes it possible to compare alternatives that are supposed to perform better such as Guiraud's corrected Type Token Ratio (GTTR) and Caroll's corrected Type Token Ratio (CTTR) that were proposed to solve a bias with Type Token Ratio (TTR). Besides using features that were used in the literature, this paper proposes new features such as Continuous Lexical Sophistication (CLS), Verb Tenses, Ngram profiles, and word embedding based coherence measurements. To compare the relevance of all these features, the effect size omega squared ($\omega^2$), was used instead of the traditional one-way Analysis of Variance ANOVA's F-test, as it is easier to interpret. Given the large number of implemented features, different experiments with three features ranking and selection methods were conducted besides omega squared. To evaluate the scalability of the proposed system to a real-world application, the system is trained and tested respectively on different data sets of different types.

The modern work about automatic text classification by linguistic complexity is related to traditional works about the readability of texts, which is about measuring how easy it is for a reader to understand a text. Many readability formulas were proposed within different contexts (e.g. readability of military documents by soldiers). Typically, these formulas were applied manually. In this paper, eight of these formulas were used as classification features and their effectiveness is compared individually among each other first and then with the related linguistic features.

According to the No Free Lunch Theorem for Optimization, formulated by Wolpert and Macready (1997), any pair of optimization algorithms is equivalent when evaluated across all possible problems. Applied to the Machine Learning (ML) algorithms, this theorem implies that there is no perfect ML algorithm that works best for all the problems. Hence, this paper compares five state-of-the-art ML algorithms: Random Forest, Multilayer Perceptron, Logistic Regression, Bagging, and Adaptive Boosting in terms of performance. Likewise, it compares the training times, regarding the number of selected features, as this time is a well-known factor to affect these algorithms' performances [Aggarwal and Zhai, 2012], [Kowsari et al., 2019].

This paper is organized as follows. In Section one, a survey of the related work is presented. In section two, the methodology is described; in section three, the used data is presented. In the sections four to nine, the adopted linguistic and psycholinguistic features are respectively described and their contributions to text complexity are evaluated using the Omega squared measure. Section ten is about the eight adopted readability formulas and their relations to the linguistic features. Section eleven describes the evaluation of different approaches to text classification by complexity and how they scale to a real-world application. Section twelve discusses the results obtained in section eleven through the patterns of classification errors. Finally, section thirteen concludes the paper and presents the future work.

## II RELATED WORK

Although some of the earliest works about text classification by linguistic complexity go back to the 19[th] century, this subject still attracts the attention of the community. Aside from ESL, many applications like first language education, stylometry, language acquisition, measurement of the virality of a post on a social network (see for example [Azpeitia et al., 2018]), and dementia diagnosis benefit from classifying the text's linguistic complexity, that is sometimes called text leveling, or text readability measurement. Therefore, there is a rich literature covering this subject from different disciplinary points of view.

In the area of language acquisition and psycholinguistics, Scarborough's Index of Productive Syntax (IPSyn) was the one that attracted much of the attention of the Natural Language Processing community [Scarborough, 1990]. IPSyn is a grammatical measure designed to show the individual differences in the acquisition of syntax. It covers sixty syntactic structures organized into four groups: noun phrases (N), verb phrases (V), questions and negations (Q), and sentence structures (S). IPSyn was implemented to score children's production using Charniak's statistical parser [Hassanali et al., 2013], [Sagae et al., 2005]. Within the framework of automatic scoring of children production, Ramirez et al., (2013) proposed some statistical measures based on word class n-grams. Lexicon was also extensively studied in relation to language acquisition, especially aspects like lexical density and diversity [Richards and Malvern, 1997], [Johansson, 2008]. Finally, [Gierut, 2007] showed that phonological features of syllabic structure or the usage of some phonemes indicate the language acquisition stage.

Within the framework of foreign language teaching, [Zampa and Lemaire, 2002], [Kurdi, 2011] adopted Latent Semantic Analysis (LSA) to recommend appropriate reading material based on the relationships between the lexicon of the texts read by the learner and the existing text candidates. In addition, several researchers have focused on the automatic assessment of ESL learners' written essays. Developed in the mid-nineties by the Educational Testing Service (ETS), e-rater combines syntactic criteria with discourse to detect and score abrupt shifts in topicality [Burstein et al., 1998]. Another system that is worth mentioning is the Intelligent Essay Assessor. It evaluates the content of essays based on Latent Semantic Analysis (LSA). It also assesses the syntactic structures and the style based on statistical measures [Foltz et al., 1999]. Besides, different works focused on specific areas of linguistic complexity. Lexicon is a key area that attracted researchers. Lexical tightness [Flor et al., 2013] and lexical focus [Kurdi, 2019] were proposed to measure the semantic proximity of the words used in a given text. These measures are more suited for content classification in open texts and are not suited to applications where all the texts are of the same type (e.g. journalistic texts).

Traditionally, grammar or more recently syntax has long been a central component of foreign language learning. This centrality led many previous studies to focus on syntactic complexity measures and their relation to foreign language proficiency. For example, [Wolfe-Quintero et al., 1998] conducted a meta-study that examined 39 works about writing development covering more than a hundred writing proficiency indicators. Ortega evaluated the usage of syntactic measures as evidence of writing development [Ortega, 2003]. Xiaofei Lu implemented a system for automatic analysis of syntactic complexity in second language writing [Lu, 2010] that uses 14 syntactic features selected among those presented in [Ortega, 2003] and [Wolfe-Qiuntero 199]. To automatically score non-native speech, Chen and Zechner (2011) collected and implemented a set of 17 key features, based on human-rated learners' transcriptions. They tested the correlation of five different models that cover each a different set of features and got encouraging results.

CohMetrix was one of the first computational tools that was created to measure the cohesion of the texts and the coherence of their "mental representation" besides other areas such as syntax and lexicon [Graesser et al., 2004]. The following criteria are adopted in CohMetrix: Co-Referential Cohesion, LSA cohesion, the relative frequency of connectives, and situation model. Co-Referential Cohesion is about content word overlap, noun overlap, argument overlap, and stem overlap. LSA cohesion measures the semantic overlap between sentences or between paragraphs. Finally, the situation model is based on the work proposed by [Zwaan and Radvansky, 1998] and provides measures of causal, temporal, and intention cohesion to account for the breadth of situation model cohesion. [Crossley et al., 2011] compared CohMetrix to two traditional readability formulas (Flesch-Kincaid Grade Level and Flesch Reading Ease) on a corpus of simplified news for L2 readers and found out that CohMetrix performed significantly better. CohMetrix relies heavily on LSA for capturing coherence and cohesion measures, which despite its advantages, suffers from well-known limitations with capturing polysemy. Besides, LSA is based on the Bag Of Words (BOW) approach where a text is represented as an unordered collection of words, unlike the newer word embedding models that capture the context of occurrence of the words. Davoodi and Kosseim (2016) investigated the role of discourse using the data set generated by Pitler and Nenkova (2009) that is made of 30 articles from the Penn

Discourse Tree Bank. This dataset comes with hand annotation for both its discourse structure and complexity. Davoodi and Kosseim examined six discourse-related features, among which four about coherence and two about cohesion. The cohesion relations are about pairs or triplets relations that may or may not be marked with a discourse marker. The cohesion features are about the number of pronouns per sentence and the number of articles per sentence. The results showed that discourse features play an important role in the classification of text complexity. The main limitation of this kind of works is that it requires data that is labeled with discursive annotations, which is hard to find especially for large-scale applications.

Different works about automatic text classification within different application contexts covered different partial combinations of linguistic features like [Davoodi and Kosseim, 2016], [Feng et al., 2010], [Vajjala and Meurers, 2013], [Xia et al., 2016], [Kurdi, 2017a], and [Balyan et al., 2018]. The results of these studies differed considerably as they did not use the same data set and did not target the same task. For example, Xia's best model reported an accuracy (ACC[1]) of 0.79, while Vajjala reported, in a series of binary classification tasks, accuracies ranging between 0.93-0.97.

Deep learning is increasingly dominating the research in NLP. Some recent work tried to apply deep learning techniques to sentence complexity classification using Long Short-Term Memory (LSTM) Networks [Boscoa et al., 2018] or text complexity classification using Gated Recurrent Units (GRU) [Nadeem and Ostendorf, 2018]. Both approaches take word embedding as an input. Despite its success in many NLP areas, deep learning suffers from a serious shortcoming. Using word embedding, which are vector representations of the actual words, means that the classification depends on the vocabulary sequences used within the corpus. This usage limits considerably the scalability of this approach, as it will not work well when facing a new text whose subject and word sequences differ from the ones covered in the training corpus.

Moreover, different models of text mining were proposed within the context of educational systems. For example, within the context of a system about the analysis of student's reviews of teacher's performance, Esparza et al. (2017) proposed an architecture, where text processing is done in three steps. First, students' comments are extracted and preprocessed. Once students' comments are extracted, they are labelled by hand into positive, negative, and neutral using a numeric range between -2 and +2. Second, after collecting and labelling the texts, feature extraction and selection is performed. Given that the task consisted of sentiment analysis, the features consisted of the text's vocabulary. Finally, the third stage consists of text classification using the SVM. [Medrano et al., 2014] proposed a more distributed architecture of text mining applied to consumption pattern identification in twitter. Three processing steps were proposed. First, relevant data acquisition from twitter as JSON document stored in MongoDB, an open source and document oriented database. Filtering the data is based on two approaches: the REST (REpresentational State Transfer) API and streaming API. The data stored in MongoDB is then accessed with a python library called urllib[2].

To summarize, many text classification works in the literature were applied to different tasks but not to text classification within ESL. Besides, the works about text classification by linguistic complexity in ESL suffer from the following limitations. Some of these works relied on a small set of linguistic features, making it hard to compare the effectiveness of different sets of features belonging to different linguistic levels in text classification by linguistic complexity. Others relied on the actual sequences of words reducing considerably the scalability. For discourse features, some works relied on old techniques such as LSA, while others required hand-annotated data. The evaluation of all the presented works was conducted using different traditional types of data splits (e.g. 70/30 split or the k-fold approach). Such splits mean that no real tests of scalability to a real-world approach using a different type of data sets for the evaluation.

---

[1] Accuracy = (TP+TN)/(TP+FP+FN+TN), where TP=True Positive, TN=True Negative, FP= False Positive, and FN=False Negative.

**III METHODOLOGY**

The main aim of this paper is to build a classifier that can distinguish between texts based on the complexity of their linguistic form. To do so, one of the most important steps consists of identifying the key features to be extracted from the text (figure 1). After a comprehensive survey of the features that were proposed in the literature, 118 features were extracted from each document. These features are distributed around five linguistic areas besides 12 psychological features and 7 readability formulas. This means that every document $D_i$ in the corpus is converted into a vector $V_i$ that has the following layout:

$V_i = < \Phi_1^{Phon}, .., \Phi_3^{Phon}, \Phi_4^{Morph}, ..., \Phi_{23}^{Morph}, \Phi_{24}^{Lex}, ..., \Phi_{35}^{Lex}, \Phi_{36}^{Synt}, ..., \Phi_{87}^{Synt}, \Phi_{88}^{Disc}, ... , \Phi_{99}^{Disc}, \Phi_{100}^{psych}, ..., \Phi_{111}^{psych}, \Phi_{112}^{READ}, ..., \Phi_{118}^{READ} >.$

The features of the same linguistic area will be ranked using the Anova's effect size omega squared ($\omega^2$). ANOVA's F-test is a common method for feature selection (see [Omer et al., 2014] and [Balyan et al., 2018] for example). The F-test is also available from popular toolkits and libraries such as Weka, Orange, and ScikitLearn. Given the difficulty of interpretation of the F-test, the effect size omega squared ($\omega^2$), which is proportional to the F-test, is used instead[2]. This choice is motivated by three reasons. First, the effect size's values are located between 0-1, which makes them more intuitive and easier to understand. Second, the existence of frameworks to interpret omega squared[3] makes the judgment about the quality of the features less subjective. According to the adopted framework, the omega squared scores that are smaller than 0.06 are considered weak, while the scores that are larger or equal to 0.06 but smaller than 0.14 are moderate. Finally, the omega squared scores that are larger or equal to 0.14 are considered as strong. Throughout this paper, the bars of the charts will be color-coded: red will represent the weak omega squared, blue will represent moderate omega squared, while green will be used to represent the strong omega squared. The p-value will be provided as an additional indication of the significance of the features.

At the global level, the adopted architecture is like the one presented in [Esparza et al., 2017]. To rank the features across all the linguistic areas, three additional feature selection methods are used: SVM, Correlation, and ReliefF.

---

[2] $\omega^2$ is an unbiased version of $\eta^2$, another effect size commonly used with ANOVA.

[3] https://easystats.github.io/parameters/reference/eta_squared.html (all the links provided in this paper are accessed for the last time in September 2019)

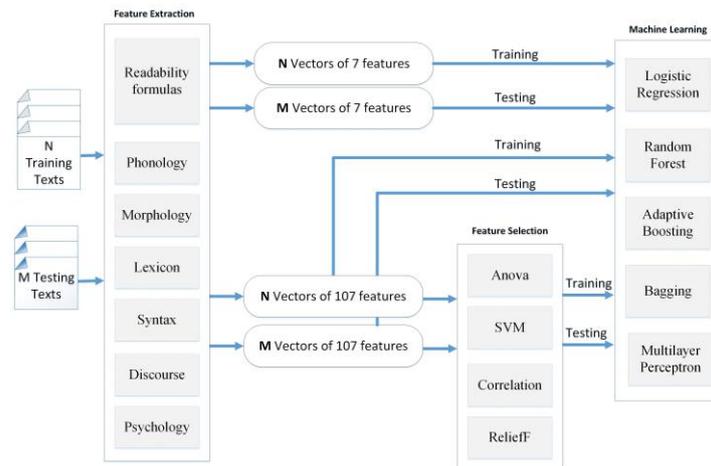

Figure 1. Data Flow diagram of the process of feature extraction, training, and testing of the ML algorithms.

After preliminary experiments with Support Vector Machine (SVM), Decision Trees, and Naïve Bayes Networks, the following ML algorithms were adopted in this study: two simple algorithms, Logistic Regression and Multilayer Perceptron, and three ensemble algorithms, Random Forests, Adaptive Boosting, and Bagging.

Given that the readability formulas are made of linguistic features such as sentence and word complexity, they will be used separately. This allows making a comparison between the classifications of the documents with the linguistic features on the one hand, and with the readability formulas on the other.

**IV CORPUS**

A corpus of (6171 total) texts of English, which are edited specifically for ESL learners, is collected from six free professional websites[4]. This corpus will be referred to by ESL Texts in Levels Corpus (ESLTL). The texts provided on these websites are organized by three levels of difficulty: 1, 2, and 3. These levels correspond respectively to A2, B1, and B2 in the Common European Framework of Reference for Languages [Council of Europe, 2011].

4964 texts in the corpus come from the *News in levels* website, which gets updated daily. Typically, on this website, the same topic is addressed within three texts of three different linguistic levels. This organization helps neutralize the bias related to the content of the texts.

Using different corpora is challenging because the standards for defining the levels of the texts may be different. However, this variation is important to show results that are not dependent on a single source of data.

The distribution of the texts over the levels is as follows: 2058 texts categorized as level one, 2085 level two texts, and 2028 texts of level three. Texts of higher levels are usually larger. This makes the collections of inferior levels smaller in size despite their larger numbers. The overall size of the corpus is 5.5 MB.

---

[4] The texts are collected from the following websites: http://www.newsinlevels.com/, https://breakingnewsenglish.com, http://learnenglishteens.britishcouncil.org/study-break/easy-reading, http://linguapress.com/inter.htm, http://www.ngllife.com/content/reading-texts-word, http://www.fortheteachers.org/Reading_Resources,
https://www.rong-chang.com (validation)

| Level | Number of texts | Mean text length in words | σ | Mean text length in sentences | σ |
|---|---|---|---|---|---|
| 1 | 2058 | 112.57 | 104.83 | 12.65 | 10.8 |
| 2 | 2085 | 129.17 | 108.4 | 9.24 | 8.09 |
| 3 | 2028 | 190.9 | 95.36 | 9.81 | 6.0 |
| Total | 6171 | 143.92 | 108.4 | 10.56 | 8.66 |

Table 1. General information about the ESLTL corpus.

As seen in table 1, the mean length in terms of words in level one and level two are close, with level two being slightly longer. The order is reversed with the maximum length in terms of sentences. This is because sentences tend to be shorter in lower level texts. Another remark about the corpus is that even though the texts of levels one and two have a smaller number of words on average, the standard deviations (σ) of these levels are high. This means that the length of the texts is not a decisive factor for the level decision. This variation in text lengths could be a source of bias. Hence, sampling and other methods will be used to avoid the bias while calculating the lexical features.

To test the scalability of the classifier, additional corpora were used. First, a corpus of 150 texts of three levels was collected: 50 texts per level (table 2). These texts were taken from three online websites specialized in children reading, the diversification of sources should avoid any bias related to a single source[5]. The three used levels target respectively the following audiences: young children, older children, and young adults. These levels correspond respectively to the levels one, two, and three in the ESLTL corpus. Given the difference in criteria between the ESLTL and this corpus, there could be some misalignments between the levels, which are an additional source of challenge to the classifier. This corpus is called the Corpus of Children Texts (CCT).

| Level | Mean text length in words | σ | Mean text length in sentences | σ |
|---|---|---|---|---|
| 1 | 169.08 | 22.55 | 22.5 | 5.41 |
| 2 | 360.84 | 77.46 | 28.22 | 10.29 |
| 3 | 413.56 | 329.05 | 24.72 | 24.04 |
| Total | 314.49 | 222.03 | 25.15 | 15.6 |

Table 2. General information about the CCT corpus.

Second, 2759 texts from the British Academic Written English Corpus (BAWE)[6] were used. The BAWE corpus is a record of texts written by proficient university-level students at the beginning of the 21st century. The covered subjects span over several areas such as agriculture, philosophy, art, and medicine. This diversity of subjects helps avoid any bias related to the topics of the papers. Given the linguistic levels of these texts, they were classified as level four. Finally, another corpus of unedited texts is extracted from the News on the Web corpus (NOW), which is made of texts from web-based newspapers and magazines from 2010 to the present time[7]. The fundamental statistics about the BAWE and NOW corpora are presented in table 3.

---

[5] https://freekidsbooks.org/reading-level/children/, https://www.eslfast.com/kidsenglish/, and http://www.magickeys.com/books/

[6] https://www.coventry.ac.uk/research/research-directories/current-projects/2015/british-academic-written-english-corpus-bawe/

[7] https://www.english-corpora.org/now/

| Corpus | Number of texts | Mean text length in words | σ | Mean text length in sentences | σ |
|---|---|---|---|---|---|
| BAWE | 2759 | 357.83 | 90.91 | 13 | 0.05 |
| NOW | 50 | 592.26 | 157.86 | 23.3 | 6.82 |

Table 3. Statistics of the BAWE and NOW corpora.

# V PHONOLOGICAL FEATURES

Phonology provides an abstract description of the sound structure of language production in terms of both segmental (phonemes or syllables) and suprasegmental phenomena such as stress and intonation (see [Kurdi, 2016] section 2.1.2 for an introduction to formal phonology). As seen in section 2, previous works have shown that there is a correlation between phonological complexity and language learnability (see [Gierut, 2007] for example).

Given that this paper is dealing with written language, focus will be made on the complexity of segmental phonological structures. Hence, two features related to segmental phonology were considered: syllables and phonemes. The idea behind these features is that a more complex text would use more words that are phonologically more complex. A third feature is added, which is the number of graphemes (or letters) to compare it with the number of phonemes.

To count the number of syllables in a word, a twofold algorithm is used. First, the CMU dictionary, available in the NLTK toolbox [Bird et al., 2009], is used. The main advantage of using a dictionary is that it is an ideal way to deal with exceptions. The CMU dictionary contains 133737 entries, and like every dictionary, it is not exhaustive. Therefore, to count the number of syllables in the words that are not covered by the CMU dictionary, an algorithm based on a count of the number of vowels, diphthongs, and triphthongs within the word is used[8]. The Omega squares of the three phonological features are provided in figure 2.

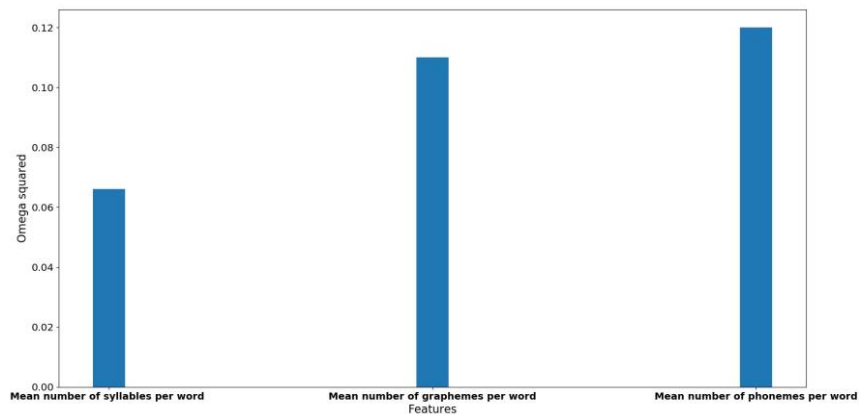

Figure 2. Omega squared of the phonological features, p<0.01 for all the features.

As seen in figure 2, the three phonological features have a significant relation with the complexity of the text. They also have moderate effect sizes. The effect sizes of the mean numbers of graphemes and phonemes per word are close to each other. This proximity is a consequence of the conceptual similarity of these two features. These two features have larger effect sizes than the mean number of syllables per word. The reason behind this difference could be the fact that the concept of the syllable is more related to spoken language production than the number of phonemes and graphemes.

Given the nature of these three phonological features, their overlap is high. The number of syllables in a word is based on the number of vowels and consonants, which is in it is turn highly related to the number of graphemes. The Pearson correlation between the mean number of the graphemes and the mean number of the syllables is [r=0.87, p<0.01, N=6171], while the Pearson correlation between the mean number of phonemes and the mean number of the syllables is [r=0.91, p<0.01, N=6171]. Finally, the Pearson correlation between the mean number of graphemes and the mean number of phonemes is [r=0.95, p<0.01, N=6171]. These high correlations confirm the tight relationship between the three phonological measures.

---

[8] https://www.howmanysyllables.com/howtocountsyllables

## VI MORPHOLOGICAL FEATURES

Morphology is about the study of the form change of the words with relation to their linguistic functions. More specifically, morphology is concerned about the way words are formed out of morphemes, the smallest meaningful linguistic units. Words being central in both spoken and written language, it is essential to explore how their complexity is related to the overall textual complexity. Hence, twenty morphological features are considered.

### 6.1 Morphological Diversity (stem diversity)

The basic assumption here is that an advanced text would use more diverse lemmas than an elementary one. A lemma is a simplified form of a word (see [Kurdi, 2016] chapter 3), for a detailed presentation of this concept). To lemmatize the words, the WordNet lemmatizer from NLTK is used. To calculate the morphological diversity, equation 1 is used.

$$\frac{\text{\# different lemmas}}{\text{total \# lemmas}} \quad (1)$$

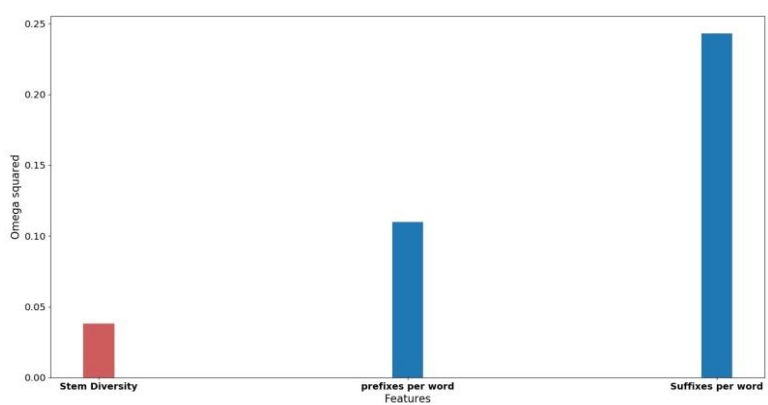

Figure 3. Omega squared of the number of affixes per word and stem diversity, for all the features P<0.01.

As seen in figure 3, the one-way ANOVA stats of this feature are significant but with a weak omega squared.

### 6.2 Mean number of prefixes and suffixes per word

This is a measure of word complexity. An advanced text would use more morphologically complex words than elementary ones. Besides, many technical words are characterized by using multiple prefixes and suffixes like in the words pre-histor-ic-al (one prefix and two suffixes) and anti-con-stitu-tion-al (two prefixes and two suffixes).

As seen in figure 3, both suffix and prefix features have significant ANOVA stats, with the mean number of suffixes per word having a higher effect size. This difference is because of many factors. It is common for a word in English to have multiple suffixes while it is not so common for a word to have multiple prefixes. In addition, suffixes play a role in determining the grammatical role of a word (e.g. *-ly* for adverb and *-tion* for a noun). These two features have a higher effect size than Morphological Diversity. A possible reason for this difference could be related to the used corpus. Despite the difference in level, the texts remain about the same subject. Thus, the amount of lemma diversity introduced is hence limited.

**6.3 Mean length of nouns, adjectives, verbs, and adverbs**
The length of some Parts-of-Speech (POS) tags within the text can be indicative of word choices, and consequently of the text complexity. Four key POS tags were considered: nouns, adjectives, verbs, and adverbs. The other POS tags such as the determiners or conjunctions were not considered since their choices are limited and they are usually implied by the grammar, not by a conscious decision of the writer.

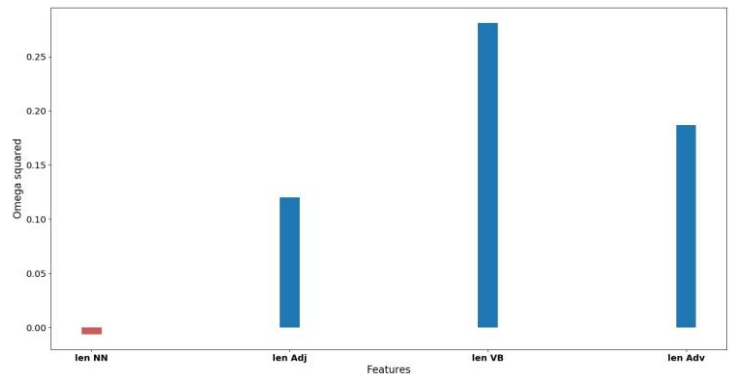

Figure 4. Omega squared of POS lengths, for all the features P<0.01.

As seen in figure 4, all the features have significant ANOVA stats, with moderate effect sizes, except for the mean length of the nouns. This is possibly motivated by the relation of the choice of the nouns to the topic of the text rather than to its linguistic form. On the other hand, the length of the verb has the strongest effect. This may be the reflection of the central role of the verb in the sentence and the big variation in verb form complexity.

**6.4 Verb tenses**
Verb tenses are one of the most fundamental elements of the grammar of a language that the learner should master. ESL textbooks classify these verbs according to complexity levels. For example, simple present and simple past are usually covered in level one while future perfect continuous is covered in level three or even later. In this study, these verb forms are considered as possible features that describe the complexity of a text.
A module that uses regular expressions of POS tags sequences to recognize 13 different verb forms is implemented. The recall of this module is 0.92, while its precision is 0.90. Most of the errors are because of issues with the POS taggers. For example, when the tagger mistakenly tags *spirit* as a verb rather than a noun. It is inserted as a simple past verb[9] [… ('in', 'IN'), ('the', 'DT'), ('Christmas', 'NNP'), ('spirit', 'VBD')]. Note that not all tagging errors involving a verb lead systematically to verb tense categorization issues. The ANOVA stats of the verb tenses are presented in table 4, along with the percentages of the text in which every tense is used. The information is coded with the same colors as in the charts.

---

[9] For a presentation of the Penn tagset used in the tagged example, please refer to the following link: https://www.ling.upenn.edu/courses/Fall_2003/ling001/penn_treebank_pos.html

As shown in table 4, we could see that only the simple future and future perfect continuous tenses are not statistically significant. This is partly because the simple future tense occurrences are evenly distributed across the levels and therefore, they do not play a distinctive role. As for the future perfect continuous, it was not observed in any of the texts, probably because it is too advanced for the adopted range of levels. Besides, simple present is the only tense with a strong effect size. This effect size results from high usage frequency overall the texts and dominance in the lower level texts, where other tenses are much less frequently used. Despite its high frequency, simple past has a moderate effect size as it is distributed over the three levels with a tendency to be more common in levels three and two. The present perfect and past perfect tenses have a mid-range frequency as each occurs in near the third of the texts, which explains their moderate effect sizes. Present continuous and past continuous have also moderate effect sizes because of their moderate frequencies. They are also relatively more common in levels three and two than in level one. Five tenses have significant ANOVA stats but weak effect sizes: infinitive, present participle, present continuous, future continuous, past continuous, future perfect, and present perfect continuous. Despite its frequent usage, infinitive has a low effect size, as it is evenly used in the three levels because of its functional role. The present participle is used in about 10% of the texts and the difference in the frequency of usage among the levels is small 34, 25, and 40 for the levels one, two, and three respectively. The other three tenses with weak effect sizes are future perfect, future continuous, and present perfect continuous. They have very low frequencies of usage ranging between 1-4% of the text.

| Tense | P value | $\omega^2$ | % of texts |
|---|---|---|---|
| simple present | <0.01 | 0.409 | 95.28 |
| simple past | <0.01 | 0.324 | 79.48 |
| present perfect | <0.01 | 0.164 | 27.02 |
| present participle | <0.01 | 0.002 | 10.16 |
| present continuous | 0.013 | 0.047 | 15.29 |
| simple future | 1 | 0 | 25.84 |
| future perfect | <0.01 | 0.011 | 4.13 |
| future continuous | <0.01 | 0.002 | 1.05 |
| past continuous | <0.01 | 0.047 | 16.26 |
| present perfect continuous | <0.01 | 0.019 | 3.06 |
| future perfect continuous | na | na | 0 |
| past perfect | <0.01 | 0.143 | 29.94 |
| Infinitive | <0.01 | 0.057 | 78.33 |

Table 4. Omega squared, p-values, and frequency of usage for the verb tenses.

**VII LEXICAL FEATURES**

Lexicon is considered in this section from a semantic point of view. It is well-known that the text meaning is a function of the meaning of the individual words that are making it up (see for example [Kurdi, 2017b] section 1.1). Thus, twelve lexical features are examined here.

**7.1 Lexical Density (LEXdens)**
Originally defined by [Ure, 1971], lexical density is the ratio of the lexical words divided by the total number of words in the text. Since lexical words are the only words that convey information, lexical density is also a measure of the information density of a text [Laufer and Nation, 1995]. Lexical words being words that are open class words, whose number is theoretically non-limited, as opposed to closed class words whose number is limited in languages. The closed class words typically include grammatical words. The obvious issue here is that this broad definition is not enough for an implementation: there is no consensus in the literature about the categories to include in each class. For example, [O'Loughlin, 1995] considered all adverbs of time, manner, and place as lexical. While [Engber, 1995] and [Lu, 2012] include in the lexical class the categories nouns, adjectives, verbs (excluding modal verbs, auxiliary verbs like be and have), adverbs with an adjective base like fast, and those formed by attaching the *-ly* suffix like particularly.

To test the different combinations, two versions of lexical densities were implemented. In the first version, the categories verbs, including modals, adverbs, including comparative and superlative adverbs, adjectives, including comparative and superlative adjectives, gerund, present participle, nouns, and proper nouns are considered as lexical. In the second version, all the common and proper nouns, adjectives, including comparative and superlative adjectives, as well as comparative and superlative adverbs are considered as lexical words. The verbs are considered as lexical except the modals *have* and *be*. Adverbs are considered as lexical when they end with *-ly* or when they have the same form as an adjective like *half*, *late*, or *low*.

As shown in figure 5, the ANOVA stats of these two versions are not significant. An interpretation of this result is that the information density is the same in the three levels, as they are about language education. Therefore, every text tries to focus on a few words.

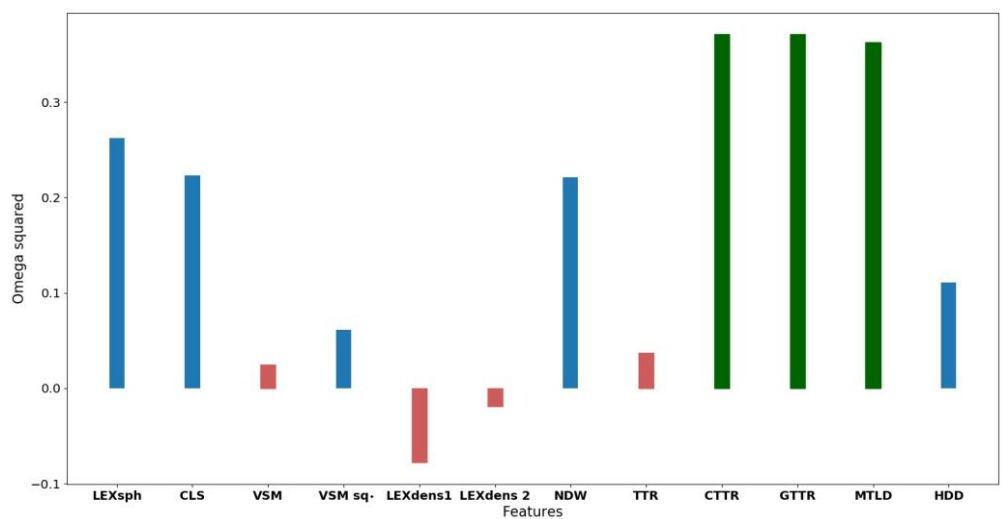

Figure 5. Omega squared of the lexical features, $p<0.01$ for all the features except for LEXdens1 and LEXdens 2, where it is one in both cases.

### 7.2 Lexical sophistication (LEXsph)

Lexical sophistication is also called lexical rareness [Read, 2000], [Lu, 2012] or Basic Lexical Sophistication. It is calculated as the ratio of the number of sophisticated words to the number of lexical words. [Hyltenstam, 1988] considers as sophisticated the words beyond 7000 most frequent Swedish words.

Given the above consideration, three versions of the Lexical sophistication are implemented. $LEX_{sph}$ is a basic version of lexical sophistication. A word is considered as sophisticated if its frequency rank is over 3000. The frequency of every word is obtained from the Word Frequency Data (WFD)[10], a freely available list of the 5000 most frequent words in English that is calculated based on the Contemporary American English Corpus [Davies, 2009]. The following stats can indicate about the coverage of WFD. 77% of the words in the level one texts are covered in the WFD, but only 66% of the words in the third level are covered in it. Furthermore, 71.50% of the texts of level two are covered. The words that are not in the WFD are given the same frequency value. This value is lower than the lowest score in the list. Words are stemmed using the Porter stemmer. Stemming helps match the singular and the plural of the regular forms (e.g. book, books). However, it does not eliminate the irregular forms such as *make* and *made*. This fits perfectly with the purpose of this score as a learner of English is supposed to know the simple plural forms of nouns but not necessarily the irregular forms of some nouns and verbs. The ANOVA stats of this feature, presented in figure 5, are significant, but its effect size is moderate. A variant of the lexical sophistication, the Continuous Lexical Sophistication (CLS), is proposed. CLS is calculated according to equation 2, based on the words that remain after filtering stop words to avoid the bias introduced by grammatical words such as determiners, copulas, and adverbs of degree. Like with $LEX_{sph}$, words are also stemmed here.

---

[10] http://www.wordfrequency.info/free.asp

$$CLS = \frac{\sum_{i=1}^{n} freq(w_i)}{n} \quad (2)$$

In equation 2, *n* is the number of words in the text that remains after the filtering.

As shown in figure 5, the ANOVA stats of CLS are significant and its effect size is also moderate.

A related measure was proposed by Harley and King (1989): the verb sophistication measure (VSM). It is calculated as the ratio of the number of sophisticated verbs to the total number of verbs (equation 3).

$$VSM = \frac{\# \text{ sophisticated verbs}}{\# \text{ verbs}} \quad (3)$$

Sophisticated verbs are defined as the verbs outside of the list of the most frequent verbs. Harley and King used two lists with respectively 20 and 200 verbs in two different studies. In both studies, they reported a significant difference between native and non-native writers. Wolfe-Quintero et al. (1998) proposed a modified version of this measure. They proposed to use the square to reduce the sample size effect (see equation 4).

$$VSM\ Sq. = \frac{\# \text{ sophisticated verbs}}{\sqrt{2 * \# \text{ verbs}}} \quad (4)$$

In this paper, the list of the 330 most frequent verbs[11] in English of the McMillan Dictionary was used. To find the uninflected form of a verb, the verb conjugation module, provided within the Pattern.en toolbox[12], is used. The ANOVA stats of the two variants, presented in figure 5, are significant. The effect size of VSM is small, while the effect size of VSM sq. is moderate.

### 7.3 Number of Different Words (NDW)

Lexical Diversity (LD) or lexical variation is a measure of the richness of the lexical forms used in a text. The intuitive way to account for LD is the count of the different lexical forms in the text, this is called the Number of Different Words (NDW). NDW was used in areas like language acquisition [Klee, 1992]. The obvious issue with adopting NDW in a text is the bias introduced by the text length. The longer the text, the bigger is the chance of observing different lexical forms. The desired diversity measure should provide an estimation independent of the size of the text. As shown in figure 5, despite its bias toward the text size, the NDW has a moderate effect size. This shows that this measure can be practically useful despite its theoretical limitation.

---

[11] http://www.acme2k.co.uk/acme/3star%20verbs.htm

[12] http://www.clips.ua.ac.be/pages/pattern-en

## 7.4 Type Token Ratio (TTR) and its transformations

Type Token Ratio (TTR) is an extended version of NDW, as it is the ratio of the Number of Different Words (NDW) to the total number of words $n$ [Templin, 1957] (equation 5).

$$TTR = \frac{NDW}{n} * 100 \qquad (5)$$

Although NDW is divided by the size of the text, it was shown that TTR is still biased toward the size of the text, since the ratio decreases as $n$ increases [Hess et al., 1986], [Richards, 1987], [Arnaud, 1992], (see [Malvern et al., 2004] for a detailed discussion). To compensate the size of the text and hence turn TTR into a constant over the whole text, several mathematical transformations of TTR were attempted. Guiraud's corrected TTR (GTTR) [Guiraud, 1960] is one of these transformations (equation 6).

$$GTTR = \frac{NDW}{\sqrt{n}} * 100 = TTR * \sqrt{n} \qquad (6)$$

Carroll (1964) proposes another transformation: Caroll's corrected TTR (CTTR). As shown in equation 7, it is like Guiraud's transformation, since we are multiplying TTR by the square of $n$, the number of words, over 2.

$$CTTR = \frac{NDW}{\sqrt{n*2}} * 100 = TTR * \sqrt{\frac{n}{2}} \qquad (7)$$

As shown in figure 5, TTR, GTTR, and CTTR have significant ANOVA stats. GTTR and CTTR have the same strong effect size, which is about ten times larger than the one of TTR. This confirms the theoretical advantage of these two transformations over the original TTR.

## 7.5 The Measure of Textual Lexical Diversity (MTLD)

The Measure of Textual Lexical Diversity (MTLD) is another way to account for lexical diversity. It is designed to reduce the effect of the text length. MTLD is computed as the mean length of sequential word strings in a text that maintains a given TTR value. During the computing process, each word of the text is evaluated sequentially for its TTR (see [McCarthy and Jarvis, 2010] for a detailed example).

According to a study on segments of spoken texts, produced by 20 intermediate non-native English speakers, MTLD is less affected by text length than TTR and GTTR, if it is used with texts of at least 100 tokens [Koizumi, 2012]. The minimum text size is set to 30 in this paper, because there are some texts from levels one and two whose lengths are smaller than this threshold (see table 2). As shown in figure 5, the one-way ANOVA stats are significant with a strong effect size.

## 7.6 HD-D

The *D* measure is another approach developed to calculate the lexical diversity in a way that is supposed to be independent of the length of the text. Proposed by Brian Richards and David Malvern, the *D* measure is based on the predicted decrease of the TTR, as the size of the text increases [Richards and Malvern, 1997]. This mathematical curve is compared with actual data from the MJ and the Lancaster–Oslo–Bergen corpora (see [McCarthy and Jarvis, 2010] and [Johansson, 2008] for more details). The *D* measure is available in the Child Language Analysis (CLAN) software, under the name VocD [MacWhinney, 2000]. Besides, McCarthy and Jarvis showed that Vocd-D is a complex approximation of the hypergeometric distribution, and to show this, they proposed an index that they called HD-D [McCarthy and Jarvis, 2010]. The hypergeometric distribution being the probability of drawing a certain number of tokens of a specific type from a text sample of a certain size. As shown in figure 5, this feature has significant ANOVA stats and a moderate effect size.

# VIII SYNTACTIC FEATURES

The syntax is a key indicator of text complexity. Sentences with complex structures are harder to understand than sentences with simple structures even for native speakers. For non-natives, the issue is that some of these structures are not even known by the learner. To extract the syntactic features, two sets of freely available tools are used. For parsing, the Stanford Parser is used [Klein et al., 2003]. Some tools from NLTK are also used, such as the sentence tokenizer and the POS tagger.

## 8.1 Phrase level

A phrase is a group of one word or more which plays the role of a syntactic constituent of the sentence. A phrase is a syntactic unit and is not defined according to any semantic constraints. It is a fundamental unit within the American Structuralist Syntax approach. Phrases are typically categorized following their central word or head. However, linguistic theories are not unanimous about the technical definition of the phrase. In this paper, phrases are extracted from parse trees got with the Stanford Parser[13].

To examine the role of the phrase level in the textual complexity, different aspects are considered. First, the total number of phrases in the text (#XP) is examined. Despite its moderate effect size (figure 6), this feature is biased against the type and the length of the text. The ratios of specific types of phrases are calculated as the number of the specific type of phrase divided by the number of phrases. For example, the ratio of Noun Phrase (NP) is calculated as in equation 8.

$$Ratio\ NP = \frac{\#NP}{\#XP} \tag{8}$$

In equation 8, $\#NP$ is the number of noun phrases in the text and $\#XP$ is the number of phrases in the text. As shown in figure 6, Verb Phrase (VP), Adverbial Phrase (ADVP), and Adjectival Phrase (AP) ratios (respectively rVP, rADVP, and rAP) have weak effect sizes. Typically, there is one VP per sentence; this makes the number of VP not very distinctive in terms of text complexity. While Adverbial Phrases can be equally used in different levels to express time or space for example. The ratios of Prepositional Phrase (PP) and Noun Phrase (NP) (respectively rPP, rNP) have moderate effect sizes, with PP having the largest among the phrases ratios. PP can be part of complex structures such as a complex NP, made for example with over one NP, a PP, or an AP, therefore they are more distinctive in leveling the texts.

---

[13] https://nlp.stanford.edu/software/lex-parser.shtml

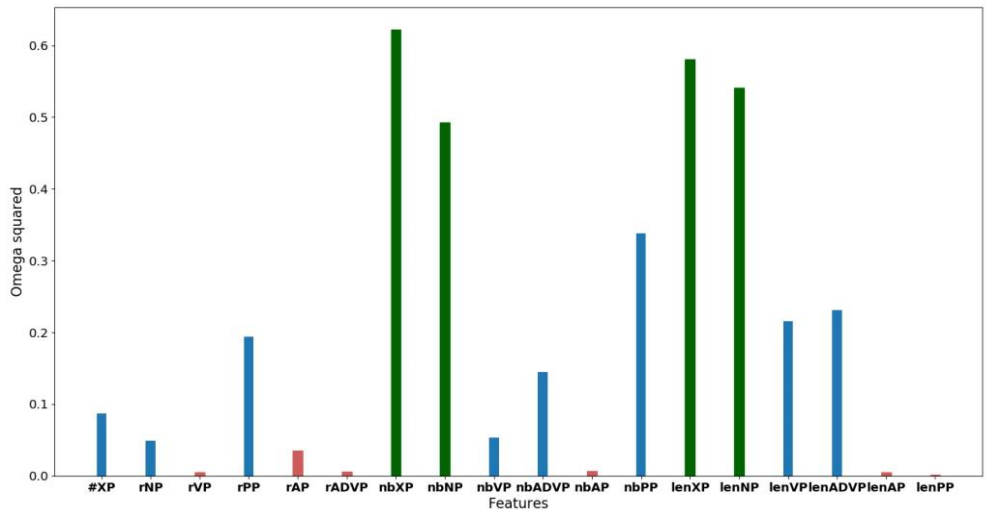

Figure 6. Omega squared for the Phrase features, for all the listed features P < 0.01.

The other group of features presented in figure 6 is the ratio of the number of phrases to the number of sentences in the text. The number of phrases and the number of noun phrases per sentence (nbXP and nbNP respectively) have strong effect sizes. The number of VP (nbVP), the number of ADVP (nbADVP) and the number of PP (nbPP) have moderate effect sizes. This confirms the limited role of the numbers of VP and its satellite the ADVP because of their small variation in the sentences across the levels. The number of adjectives per sentence (nbAP) has a weak effect size.

With the mean length of phrases, the mean lengths of the NN and XP have a strong effect size. On the other hand, the mean lengths of the adjective phrase (lenAP) and the mean length of the Prepositional Phrase (lenPP) are weak, while the mean lengths of the VP and ADVP have moderate effect sizes.

As seen in figure 6, there is a difference between the effects of features related to the same syntactic constituent. For example, the length of the ADVP has a moderate effect size while its ratio per sentence has a weak effect size. On the other hand, the number of phrases has a moderate effect size, while the number of phrases per sentence and the mean length of phrases have strong effect sizes. This shows that the explored variations of different aspects of the same syntactic constituent are relevant.

**8.2 T-unit**

Coined by [Hunt, 1965], the term T-unit typically refers to the shortest grammatical sentences. In many cases, but not all the time, a T-unit is a sentence. T-unit is made with one main clause followed by any possible clausal or non-clausal subordinates. The algorithm used in this paper to extract the T-units from the parse trees is inspired by [Lu, 2010], where T-units are searched within one sentence. Three features related to the structure of a T-unit are considered the mean numbers of VP, NP, and PP per T-unit. The ratio of complex T-units in the text (# of T-units / # of complex T-units) as well as the mean length of T-units, are used. All the T-unit features have both significant and strong effect sizes, with the mean number of PP in T-unit having the strongest effect size (figure 7).

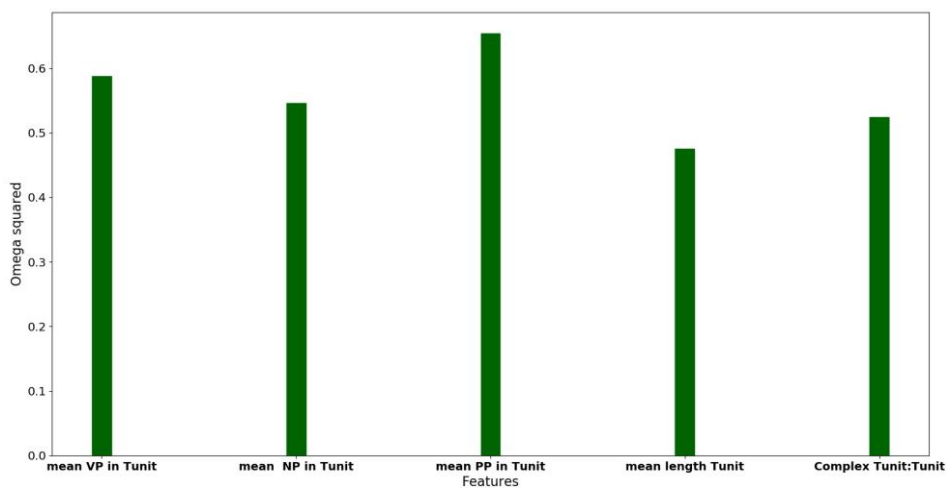

Figure 7. Omega squared of the T-unit features, for all the listed features P < 0.01.

### 8.3 Sentence

The sentences are identified using NLTK's sentence tokenizer. As seen in figure 8, all the sentence related features have significant ANOVA stats. The mean length of the sentence (meanLenS), the height of the parse tree (hghtTree), the mean number of subordinated classes (subord), and length of subordinated sentences (lenSubord) have strong effect sizes. The length of a sentence (meanLenS) is a simple but good indicator of the complexity of its structure as well as the diversity of the words used in it. The strong effect size of this feature confirms the results observed in other studies [Chen and Zechner, 2011], [Lu, 2010].

Inverted declarative sentences (invDecS) are sentences where the subject follows the tensed verb or modal. For example, in the declarative inverted sentence *are late all the products arriving tomorrow* the subject *products* occurs after the verb *are*. This feature has a moderate effect size as it is more likely to occur in level three than in level two, where it is also more likely than in level one, with a small difference between level two and level one. The length of the inverted declarative sentences (lenInvDecS) has a weak effect size.

Interrogation by inversion (invQst) is a common structure in English. For example, in the sentence *did you go to Paris?* The verb *did* occurs before the subject *you*. This structure is also shared among other European languages like French, which makes it even more accessible to beginners. This feature has a strong effect size, while its length (lenInvQst) has a weak effect size. In addition to be used in all the texts of the corpus, Wh-questions have distinctive patterns of occurrence as they are more likely to occur in level one than in level two, and in level two than in level three. Therefore, the mean number of wh-questions per sentence (whQst) has a strong effect size. The mean length of wh-questions (lenWhQst) has a moderate effect because of the limited variations of the lengths of this feature.

Coordination can apply to entire sentences or clauses, like in *He may buy a gift to his son* or *he may take him to the park*. Two features related to the coordination of sentences are examined: the mean number of the coordinated sentences (sentCord) and the mean length of the coordinated sentences (lenSCord). These two features have both weak effect sizes. This is because of the rarity of this phenomenon: it was only used in 182 texts (out of 6171 texts). Furthermore, the distribution of the occurrences of this structure across the three levels respectively is as follows: 40, 47, and 96. This means that the likelihood of occurrence is similar in level one and level two. Coordinating phrases within a sentence is another possible indicator of complexity. This can give sentences like *I want to buy a red fish and a blue bird* (two coordinated phrases) or *I want to buy a red fish, an orange cat and a blue bird* (three coordinated phrases). Phrase coordination is examined here from two points of view. The first is the mean length of the coordinated phrases (lenXPCord) and the second is the mean number of coordinated phrases (XPCord). These two features have moderate effect sizes, as seen in figure 8.

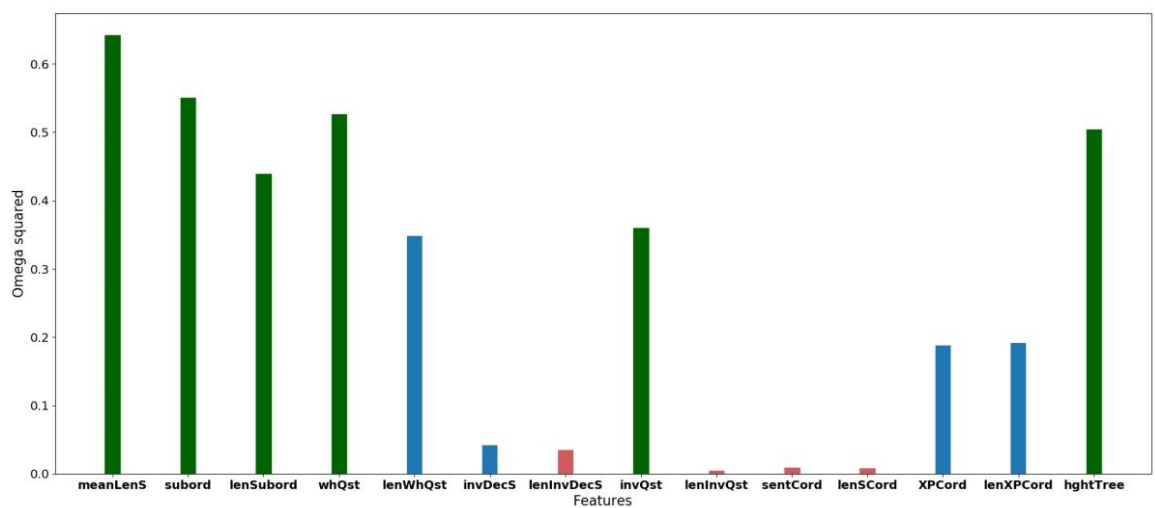

Figure 8. Omega squared of the sentence features, for all the listed features P < 0.01.

## 8.4 Ngram based features

Ngrams are a good measure of the co-occurrence of syntactic categories. They have been used extensively as language models in areas like speech recognition, information retrieval, biological sequence analysis, and data compression. Some previous studies tried to capture different aspects of syntactic complexity related to language acquisition through n-gram based models [Ramirez de la Rosa et al., 2013].

Here, the number of bi-grams, tri-grams, and four-grams of POS tags' sequences are in two different ways. The first one considers the ratios of ngrams per word and the second considers the ratio of ngrams per sentence (figure 9). An advanced text should have more diversified sequences of POS tags.

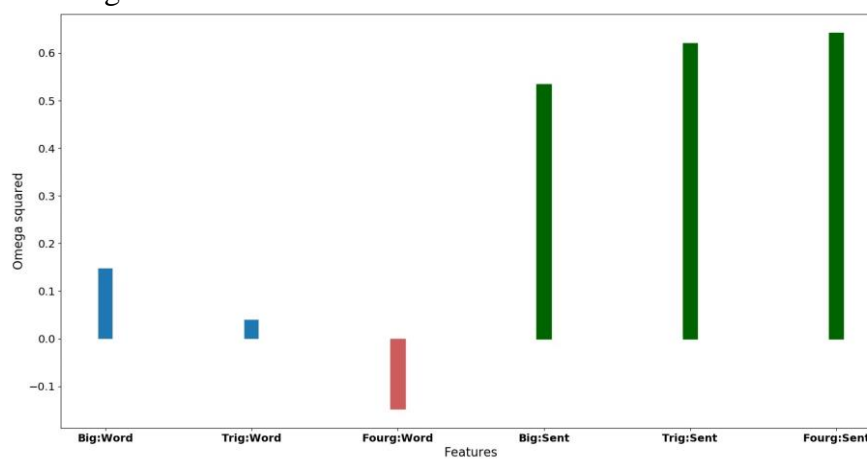

Figure 9. Effect sizes of the ngram features, p<0.01 for all the features except for fourgram per word.

As seen in figure 9, the three ratios of ngrams per sentence have strong effect sizes. This is not the case with the three ratios of ngrams per word, where the mean number of bigrams and trigrams per word have moderate effect sizes, while the mean number of fourgram per word was not significant (p=1 with a negative effect size). The above results show that the measure of sentence complexity is more significant than the general measure of complexity over the words of the entire text.

Inspired by the work of Cavnar on language identification [Cavnar and Trenkle, 1994], the other method of using ngrams consist of building ngram profiles of POS tags for every level of texts. Hence, for every level, three profiles are built: bigram, trigram, and fourgram. Three profiles are also built for each text. The profiles are stored in separate dictionaries: key-value data structures, where the keys are the ngrams and the values are the frequencies. The built dictionaries are then normalized, so the frequency of each ngram is replaced with its rank. For example, the frequency of the most frequent ngram is replaced by its rank 1, and so on.

Then, for every text, its distance with the three profiles of the same type is measured. For instance, the distances between the bigram profile of the text are measured with the three bigram profiles of the three levels respectively. The distance between the profile of a text and the profile of a level is the absolute value of the subtraction between the ranks of every ngram in the text and level profiles. More specifically, for every ngram in the text's dictionary, the absolute value of the subtraction of the rank of this ngram in the text from the rank of this ngram in the level profile is added to a delta variable, which is the actual distance between the text and the level profile (see [Cavnar and Trenkle, 1994] for more details). The omega squared of the ngram profiles are provided in figure 10.

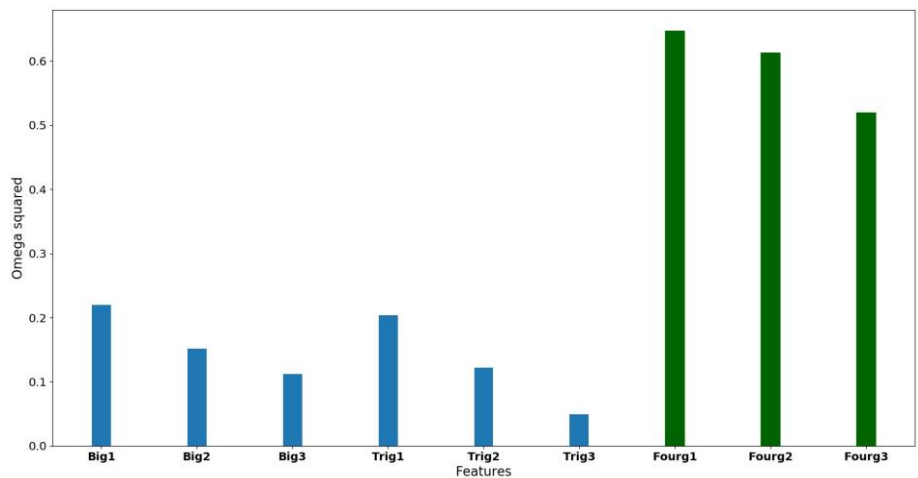

Figure 10. Omega squared for the ngram profiles features.

We can see in figure 10, the effect size decreases with the increase of the texts' levels of the model. For example, the fourgram model of level one has a larger effect size than the fourgram model of level two, which in its turn has a larger effect size than the fourgram of level three. Furthermore, the fourgram profiles are more distinctive, as they have strong effect sizes, while the trigrams and bigrams have moderate effect sizes. This can be explained by the bigger expressive power of fourgrams. The sequences of two or three POS tags are not large enough to be characteristic of the level of text.

**IX INTER-SENTENTIAL FEATURES**

As seen in section 2, Inter-sentential or discursive features play an important role in text complexity description. Several models of the text's discourse were proposed (see (Kurdi, 2017) for a review). In this paper, two aspects of discursive complexity will be covered: cohesion and coherence.

## 9.1 Cohesion and Discourse connectors

The cohesion of a text is about the explicit relationships between the discourse components that are usually clauses, and sentences. A simple but reliable way to measure the cohesion of a text is the discourse connectors. Discourse connectors, such as *and*, *therefore*, and *hence*, indicate long and elaborate sentences as well as an advanced structure of the text. Hence, discourse connectors are considered as a factor of complexity. Two cohesion features are calculated: the ratio of discourse connectors per word and the ratio of argumentative discourse connectors per word. Argumentative discourse connectors such as *hence*, *because*, and *thus*, are a subset of discourse connectors that indicate a higher level of reasoning and argumentation. To calculate the ratios of the discourse markers in the text, equation 9 is used, where *n* is the total number of words in the text.

$$connectors:words = \frac{\# Discourse\ connectors}{n}$$
$$connectors:sentences = \frac{\# Discourse\ connectors}{\# sentences}$$
(9)

Furthermore, the ratios of the discourse connectors and argumentative discourse connectors per sentence are considered. This gives four measures of cohesion (figure 11).

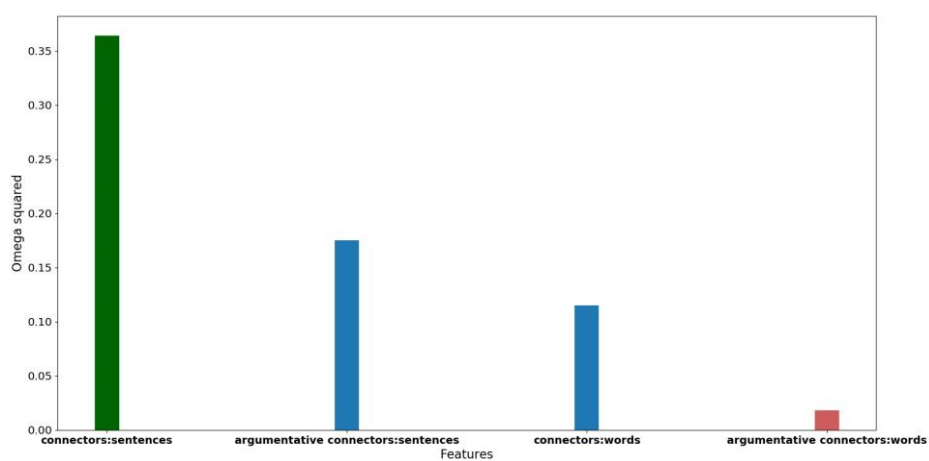

Figure 11. Omega squared of the cohesion features, pValue <.01.

As we can see in figure 11, the four cohesion features are statistically significant. The ratio of discourse connectors per word has a moderate effect size, while the ratio of argumentative discourse connectors has a weak effect size. The smaller number of occurrences can be the reason for the weak effect of the argumentative connectors: there are 47375 occurrences of discourse connectors in the corpus, with 21065 occurrences of argumentative discourse connectors. Another observation is that the sentence plays an important role at the cohesion level, since the effect sizes of the ratios per sentence are higher than the effects of the ratios per word.

## 9.2 Coherence

Coherence is about the way a text establishes semantic relations within and between sentences [Hobbs, 1979]. As seen in the literature review, some previous works relied on hand-annotated data to mark the pairs or triplets of sentences with coherence related relationships, such as cause or explication. Unfortunately, such data is not always available in the right quantity. Therefore, in this paper, coherence is measured as the mean bi-sentential distance. To have accurate distance, anaphora was replaced with their referents using the Hobbs algorithm [Hobbs, 1978].

For example, in the sentences: *The students go to school every morning.* ***They*** *enjoy riding the bus*. The anaphoric *they* in the second sentence is replaced by its antecedent *the students*.

The coherence of the text is measured using the distance between the words of every pair of contiguous sentences (equations 10 for the bi-sentential distance and 11 for the distance between a word and a sentence). A version marked with NN of some features consists of keeping the nouns only was also used. This helps test the centrality of the nouns in the sentence's semantics.

$$sim(w, S) = \frac{\sum_{j=1}^{k} dist(w, w_{S(j)})}{k} \qquad (10)$$

In equation 10, *n* is the number of nouns in the sentence $S_m$ and $w_i$ is the i$^{th}$ noun from $S_m$.

$$PsyF_i = \frac{\sum_{j=1}^{k} MRC_{PsyFi}(nn_j)}{k} \qquad (11)$$

In equation 11, *k* refers to the number of nouns in the sentence S, $w_{s(j)}$ is the j$^{th}$ noun from *S*, and *w* is a noun from the next sentence. The function *dist* uses four semantic measurements. The first is the Wu-Palmer similarity (WUP), which is based on the depth of the two senses of these words in the taxonomy and that of their most specific ancestor node. To avoid the limitations of LSA that uses the Bag Of Word (BOW) approach for building the document vectors, some additional experiments were conducted using three approaches to word and document embedding: document to vector (doc2vec) [Mikolov et al., 2013], Word to Vector (word2vec) [Le and Mikolov, 2014], and Glove [Pennington et al., 2014]. The effect sizes of the coherence features are presented in figure 13.

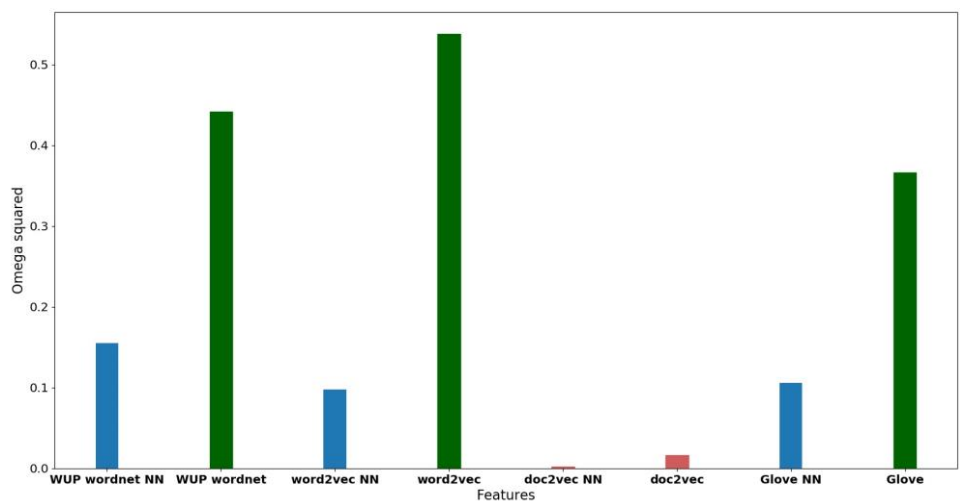

Figure 12. Omega squared of discourse coherence features, p<0.01.

As shown in figure 12, using the nouns only to measure the bi-sentential distance is not an effective strategy, as the effect sizes of the measures with NN are systematically lower than the effects of the same measures but with all the words. Another observation is that doc2vec is the only measure to have a weak effect size with all the words. Finally, word2vec with all the words has the highest effect size.

## X PSYCHOLINGUISTIC FEATURES

Several psycholinguistic measures are also covered in this paper such as Kucera-Francis Written Frequency (KFWF), Kucera-Francis number of categories (KFNCat), Kucera-Francis number of samples (KFNS) [Francis and Kucera, 1982], Thorndike-Lorge written frequency, Brown verbal frequency, Familiarity rating (famil), Concreteness rating (concrete), Imageability rating, Meaningfulness (Colorado Norms) [Nickerson, 1984], Meaningfulness (Paivio Norms), as well as the Age of Acquisition rating (Aquis age).

For every lexical item in a text $T$, these psycholinguistic features' values $PsyF$ are taken from the MRC[14] database as in equation 12. First, the list of the nouns is extracted from the text and their total number of occurrences $k$ is counted. The scores of every psychological feature, as extracted from the MRC dictionary, are added and then divided by the number of words in the text. This helps get the mean value of this feature in the text. In equation 12, $i$ is the i[th] psycholinguistic feature, $nn_j$ is the j[th] noun in the text.

$$PsyF_i = \frac{\sum_{j=1}^{k} MRC_{PsyFi}(nn_j)}{k} \qquad (12)$$

Finally, the ratio of the number of nouns that are in the MRC to the number of nouns in the text is also used as a feature.

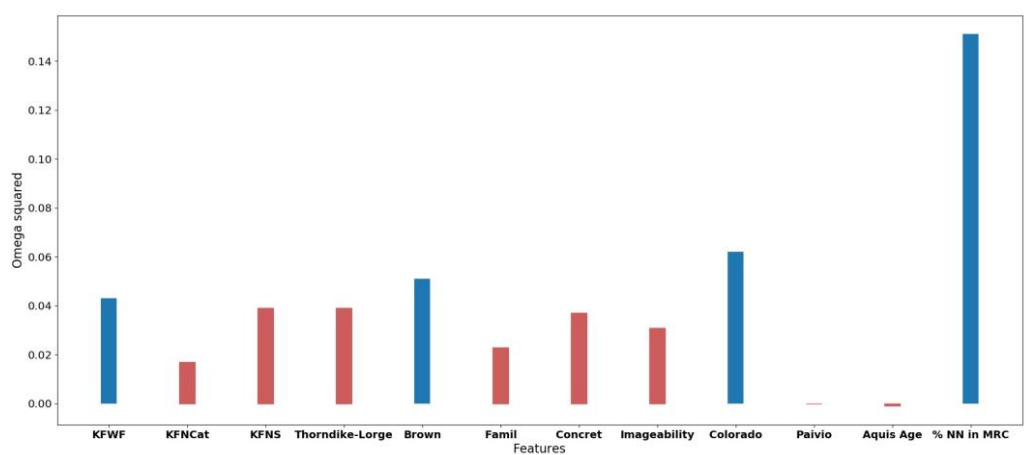

Figure 13. Omega squared of the psycholinguistic features, for all the features P<0.01, except for the Paivio norms and the age of acquisition where it was 0.094 and 1 respectively.

As seen in figure 13, the age of acquisition and the Paivio norms are the only statistically insignificant features. Besides, only four features have moderate effect sizes, while the rest of the features have weak effects sizes. The feature with the largest effect size is the ratio of words that are in MRC. A justification of these moderate to weak effects of the psycholinguistic features is that some of these features are originally designed for studies about language acquisition. Furthermore, many of these features, like familiarity, concreteness, and imageability are about the content of the words, which is similar across the three levels.

**XI READABILITY FORMULAS AS TEXT FEATURES**

Several formulas have been proposed to assess the difficulty, understandability, or readability of texts. As seen in section 2, some automatic classification works have used these formulas. The goal of this section is to compare these readability formulas and analyze their suitability as features within a classifier. These different formulas are all based on criteria that have already been explored in this paper. Therefore, the only added value with these formulas is the weighting process of every source of information. All considered formulas have statistically significant ANOVA stats (figure 14).

---

[14] http://websites.psychology.uwa.edu.au/school/MRCDatabase/uwa_mrc.htm

**11.1 Gunning's Fog index**
Developed by Robert Gunning, this formula was initially designed to estimate the number of years of formal education a person needs to understand a text after a first read [Gunning, 1952]. It is used today to decide if a text's level is suited for the targeted audience or not (equation 13).

$$0.4 \times \left(\frac{\text{\# words}}{\text{\# sentences}} + 100 \times \frac{\text{\# complex words}}{\text{\# words}}\right) \quad (13)$$

As we can see in equation 13, there are two major terms in the fog index's formula. The first is about the number of words per sentence (mean length of the sentence), which is an important indicator of text complexity, as shown in section 8.3. The second term is about the percentage of complex (or foggy) words. Complex words are words with three or more syllables. Hence, the Gunning's Fog index is based on a combination of a measure of sentence complexity with a measure of word complexity (morphology and phonology).

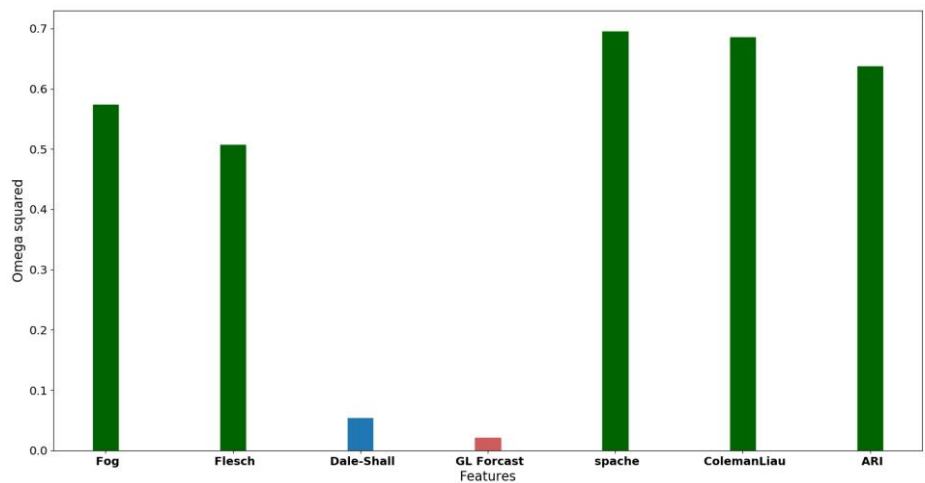

Figure 14. Omega squared of the readability formulas, for all the listed features P < .01.

As seen in figure 14, the Gunning's Fog index has a strong effect size (0.57), though its effect is smaller than the mean sentence length effect alone (which is 0.64).

**11.2 The Flesch-Kincaid formula**
Developed separately by Rudolf Flesch and Peter Kincaid, this formula was used in the US Navy and US military to assess the difficulty of texts [Flesh, 1981], [Kincaid, 1983]. They assume that the formula is based on the way humans understand written language. One of the underlying ideas in this formula is the promotion of a return to phonetics to improve writing. The Flesh-Kincaid formula is provided in equation 14.

$$Flesch - kincaid\ Score\ =\ 206.835 - \left(1.015 \times \frac{\text{\# words}}{\text{\# sentences}}\right) - \left(84.6 \times \frac{\text{\# syllables}}{\text{\# words}}\right) \quad (14)$$

Like in the Fog's Index score, there are two major terms in the Flesh-Kincaid formula. The first is about the mean number of words per sentence and the second is about the mean number of syllables per word. The Flesh-Kinaid has a strong effect size (0.50), which is smaller than the effect size of Gunning's Fog index. Here as well the effect size is smaller than the one of the mean sentence length effect alone, but it is much larger than the effect of the mean number of syllables per word (0.06).

### 11.3 Coleman-Liau Index

Designed to approximate the required US grade level to understand a text, this test was developed by Meri Coleman and T. Liau. The formula of this test is provided in equation 15, where *L* and *S* are respectively the mean numbers of letters and the mean number of sentences in the sample. In the original work, Coleman and Liau use a sample of 100 words. Given the limited size of the texts in the ESLTL corpus used in this study, samples of 27 words are used.

$$CLI = 0.0588 \times L - 0.296 \times S - 15.8 \tag{15}$$

Here as well, the score is based on two parts. The first term is related to the mean length of the word: the more characters we have within the sample, the bigger is the mean length of the words. The second is about the length of the sentences: the more sentences within the sample, the shorter is the mean length of the sentence. This measure has a strong effect size (0.685) despite the reduction of the size of the sample.

### 11.4 Spache readability formula

Designed by George Spache, this method is used to assess the readability of young children's texts up to the fourth grade [Spache, 1953]. It is initially designed to predict the grade for a text (equation 16).

$$SPACHE\ score = \left(0.121 \times \frac{\#\ words}{\#\ sentences}\right) - \left(0.082 \times \frac{\#\ unfamilar\ words}{\#\ words}\right) + 0.659 \tag{16}$$

As we can see in equation 16, the formula combines a syntactic criterion, which is the sentence's mean length, with a lexical criterion, which is the percentage of unfamiliar words. By unfamiliar words, Spache means all the words that are not part of his list, which comprises 925 words[15]. As shown in figure 14, this feature has a strong effect size (0.706), which is larger than the sentence's mean length.

### 11.5 The Dale–Chall formula

Edgar Dale and Jeanne Chall proposed a readability measure similar to the measure of Flesch-Kincaid. Instead of using the mean number of syllables per word as a measure of word complexity, they used a list of the most common words in English. They proposed a list of 3000 words that should be familiar to a large extent to children in 5$^{th}$ grade[16]. The formula is provided in equation 17.

$$DALE\text{–}CHALL\ score = 0.1579 \times \left(\frac{\#\ difficult\ words}{\#\ words} \times 100\right) + 0.0496 \times \left(\frac{\#\ words}{\#\ sentences}\right) \tag{17}$$

If the percentage of difficult words is over 5%, then one must add 3.6365 to adjust the score. This feature has a moderate effect size effect (0.05), which is smaller than the one of the Spache formula. This formula is almost identical to the Spache readability formula except for the weights and the list of difficult words. On the one hand, it gives more weight to the difficulty of the word and, on the other hand, it is possible that the choice of difficult words by Spache is more appropriate.

### 11.6 Automated Readability Index (ARI)

Like the other readability formulas, it provides a score that assesses the readability of a text. Smith (1967) developed it (equation 18). The higher the score the higher the level of the text. Like the Coleman-Liau index, it relies on the number of characters instead of the number of syllables to measure word complexity. This helps avoid any issues related to calculate the number of syllables, as this is not a trivial process.

---

[15] The Spache word list is available at this link: http://www.readabilityformulas.com/articles/spache-formula-word-list.php

[16] The Dale-Chall word list is available at this link: http://www.readabilityformulas.com/articles/dale-chall-readability-word-list.php

$$ARI\ score = 4.71 \times \left(\frac{\#\ characters}{\#\ words}\right) + 0.5 \times \left(\frac{\#\ words}{\#\ sentences}\right) - 21.43 \tag{18}$$

This feature has a strong effect size (0.64). Its effect is almost identical to the one of Coleman-Liau despite the differences in terms weighting.

### 11.7 The FORCAST Readability Formula

In 1973 John S. Caylor, Thomas G. Sticht, and J. Patrick Ford proposed the forecast readability formula within the framework of a military project whose goal was to propose a measure of the reading requirements of military specialties in the American Army. Unlike most readability formulas, the FORCAST formula focuses on the complexity of the words, thus ignoring the sentence complexity. The FORCAST formula calculates the number of monosyllabic words within a sample of 150 words. The formula is provided in equation 19, where $N$ is the number of monosyllabic words within a sample of 150 words.

$$20 - \left(\frac{N}{10}\right) \tag{19}$$

Given the limited size in the ESLTL corpus, a sample size of 27 words is used. Consequently, $N$ is divided by two instead of by ten to have a score like the original formula. The FORCAST formula has a weak effect size (0.22). This is a normal consequence of the limitation of the considered aspects of text complexity as well as the small window size that the analysis is limited to, because of the size of the texts in the ESLTL corpus.

### 11.8 Correlations of the readability formulas

As presented in the previous section, most of the readability formulas are based on similar criteria: a mixture of the measures of syntactic complexity as well as morpho-phonological complexity. In table 5, the Pearson correlations of these formulas are presented to determine their redundancy. To understand the Pearson correlations in table 5, a recapitulation of the knowledge resources used in the formulas is presented in table 6.

| Readability Formulas | Flesch-Kincaid | Dale–Chall | FORCAST | Spache | Coleman-Liau | ARI |
|---|---|---|---|---|---|---|
| Fog | -0.859 | 0.582 | 0.433 | 0.843 | 0.703 | 0.934 |
| Flesch-Kincaid | | -0.413 | -0.276 | -0.995 | -0.807 | -0.913 |
| Dale–Chall | | | 0.458 | 0.386 | 0.284 | 0.6 |
| FORCAST | | | | 0.251 | 0.199 | 0.449 |
| Spache | | | | | 0.812 | 0.901 |
| Coleman-Liau | | | | | | 0.724 |

Table 5. Pearson correlations of the readability formulas, for all the correlation N=6171, p<0.01.

The correlations between FORCAST and the other formulas range between moderate and weak[17]. This can be explained by the fact that FORCAST, unlike the other formulas, uses only one source of information, which is the percentage of monosyllabic words in a sample. The Gunning's Fog index has strong positive correlations with four other formulas: Spache, ARI, Coleman-Liau, and Dale–Chall. This suggests that the role played by the word form or phonological complexity is like the one played by the measure of word frequency of usage. The strong negative correlation with Flesch-Kincaid a result of the fact that in this last formula the two terms are subtracted from 206.83. Besides, all the formulas, excluding FORCAST, have at least one strong correlation with another formula.

---

[17] A correlation between 0-|0.30| is considered as weak, |0.30|-|0.50| is moderate, and |0.50|-|1| is strong.

| Readability Formulas | Sentence complexity | Word graphic or phonological complexity | Word frequency of usage |
|---|---|---|---|
| Fog | + | - | + |
| Flesch-Kincaid | + | + | - |
| Dale–Chall | + | - | + |
| FORCAST | - | - | - |
| Spache | + | - | + |
| Coleman-Liau | + | + | - |

Table 6. Knowledge resources used in the readability formulas.

As a general conclusion about the usage of readability formulas as features of text complexity classification, these formulas effect sizes are generally smaller than the effect sizes of their ingredient features, especially the sentence length. Even with Spache, its effect is smaller than the effects of its components: sentence complexity and lexical sophistication. Except for FORCAST, which has a weak effect size, all the formulas are highly redundant compared to each other. These formulas are redundant with their basic ingredients as well. For example, the Pearson correlation between the Gunning's Fog index and lexical sophistication is [$r$=0.718, N=6171, $p < 0.01$] and with sentence length is [$r$=0.846, N=6171, $p < 0.01$]. Both correlations are strong.

## XII EXPERIMENTS ON TEXT CLASSIFICATION AND RESULTS

Five Machine-Learning algorithms are adopted in this paper. Logistic Regression, Multilayer Perceptron, Adaptive Boosting (AdaBoost), bagging [Breiman, 1994], and Random Forest [Ho, 1995]. These algorithms were selected for their better performance after having done some experiments with other algorithms such as Decision Trees, SVM, and Naïve Bayes.

A cross-validation approach was carried with 10 folds to avoid any bias related to the partition of the data between the training and testing data sets. To measure the performance of the different algorithms, the following measures were adopted: recall, precision, and F-score (see [Kurdi, 2017b]), as well as Matthews's Correlation Coefficient[18] (MCC), and Roc Area[19]. The experiments were conducted using the Weka workbench[20].

### 12.1 Overall classification rates

The first experiment is with all the 111 features of all the linguistic areas. This experiment insights about the performance of the five adopted ML algorithms (figure 15).

---

[18] MCC= $\frac{TP \times TN - FP \times FN}{\sqrt{(TP+FP)(TP+FN)(TN+FP)(TN+FN)}}$ where TP is the number of True Positive cases, TN the number of True Negative cases, FP is the number of False Positive cases, and FN the number of False Negative cases

[19] The ROC is a probability curve that is plotted with TPR (True Positive Rate) on y-axis against the FPR (False Positive Rate) on the x-axis. Where $TPR = \frac{TP}{TP+NN}$ and $FPR = \frac{FP}{TN+FP}$

[20] https://www.cs.waikato.ac.nz/ml/weka/index.html

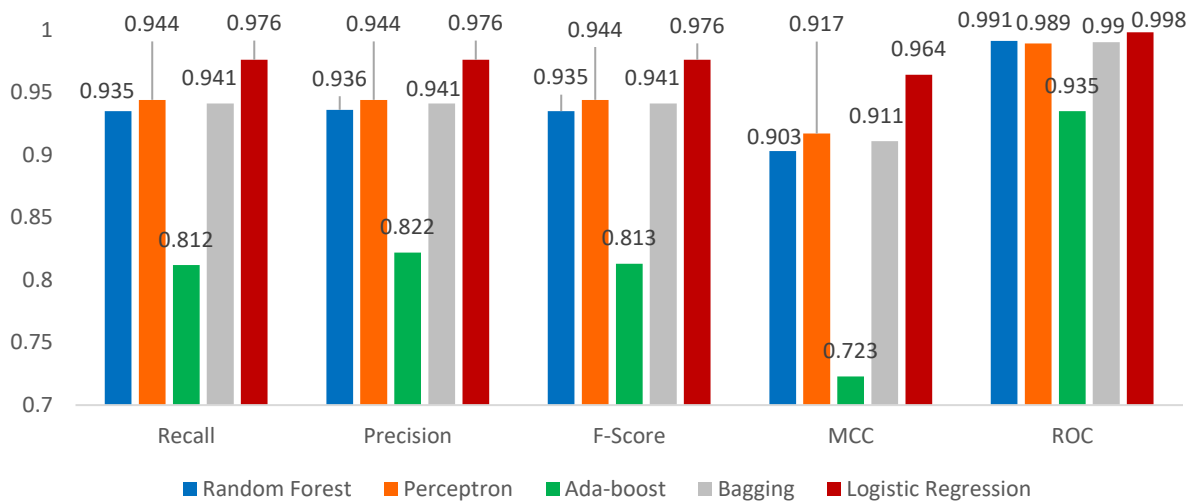

Figure 15. Classification results by linguistic complexity with all the features.

## 12.2 Scalability experiments

The results presented in figure 15 are positive. However, it is important to consider if such a classifier can handle open texts within a real-world scenario. For example, plugging the classifier into a tool to recommend a text to a student or a teacher from a large pool of texts, or even directly from the web through a search filter, are both real-world scenarios that ought to be considered. To check if such classifier scales to the real-world problem, two different supplemental evaluations with different corpora were carried.

The first evaluation was carried with the goal of testing whether the system can scale to adding an extra level of complexity, called here level four, which is about unedited texts written by proficient English writers. A combination of the ESLTL and the BAWE corpora is used with the 10-fold approach. The 2759 texts from the BAWE corpus are considered as level four. The results are presented in figure 16.

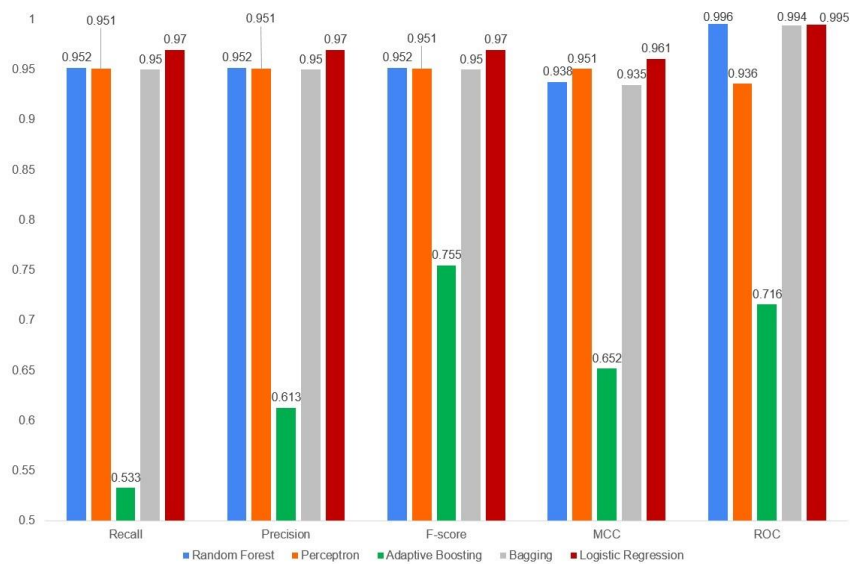

Figure 16. Classification results by linguistic complexity with all the features and the ESLTL and BAWE.

The second experiment consists of training the system on the ESLTL (for levels one, two and three) and the BAWE (for level four) corpora and use the CTT (for levels one, two and three) and NOW (for level four) corpora for the evaluation. The goal of this evaluation is to show that a classifier trained on a given type of texts can be used with different types of texts that can be found on the web. In this scenario, the system is facing the following challenges. On the one hand, the texts in the CTT corpus, being for children, are not written with the same linguistic complexity criteria as the ones of the ESLTL. Hence, a text may be considered as level two in the CTT but could be considered as level three in the ESLTL. On the other hand, the texts in the NOW corpus (level four) are of a different type than the texts from the BAWE corpus (journalistic vs. college student essays). The results of this evaluation are presented in figure 17.

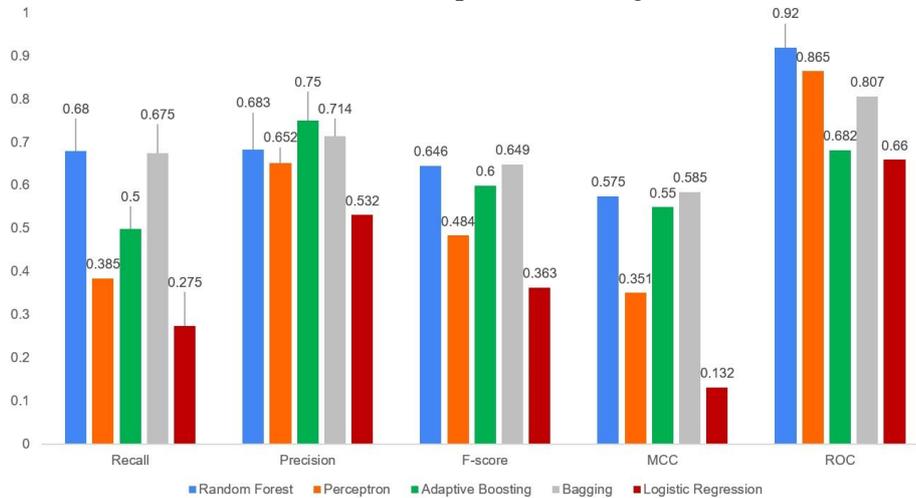

Figure 17. Scalability test results of four texts levels with the CTT and NOW corpora.

## 12.3 Automatic feature selection experiments

As seen in the previous sections, 111 linguistic features are explored in this paper, with different effect sizes. Using such a large number is computationally expensive and can lead to overfitting. The question now is which features are necessary to achieve an optimal classification rate? A rate is considered optimal if it is equal or superior to the rate got with the totality of the features. Thus, several experiments with different feature selection methods were conducted to determine which features are necessary to achieve an optimal classification rate.

A first approach of feature selection can be done with automatic methods. Hence, five different methods of feature selection were adopted. Besides ANOVA's omega squared effect size, Correlation, SVM, and ReliefF are used.

Correlation-based Feature Selection CFS was proposed in Mark Hall dissertation [Hall, 1999]. The idea with this method is to select the features with the highest correlation with the class and lowest interrelation. ReliefF extends the relief algorithm developed by [Kira and Rendell, 1992] that can handle multi-class classification applications. This algorithm consists of selecting randomly an instance (vector of features) and finds its $k$ nearest neighbors that belong to the same class (nearest hits) as well as the $k$ nearest neighbors that belong to a different class (nearest misses). It then calculates the quality estimate of the features using the same formula as relief. Support Vector Machine (SVM) is a machine-learning algorithm that can be used for feature selection. The main goal of this algorithm is to find an optimal hyperplane that can separate the vectors of different classes. Once SVM is fitted and the optimal hyperplane is found, it is possible to obtain the scores of the features.

A first experiment to test the model consists of using a gradual number of features ranked by the different feature selection methods. The goal of this experiment is to show if adding more features influences the classification rate. The result of a gradual number of features from 10 to 111 (the totality of features) is depicted in figure 18 using the ReliefF method. The x-axis represents the features grouped by 10 and ranked by reliefF. For example, 1-10 features have the highest reliefF scores; the features 11-20 have the second-highest reliefF scores, etc.

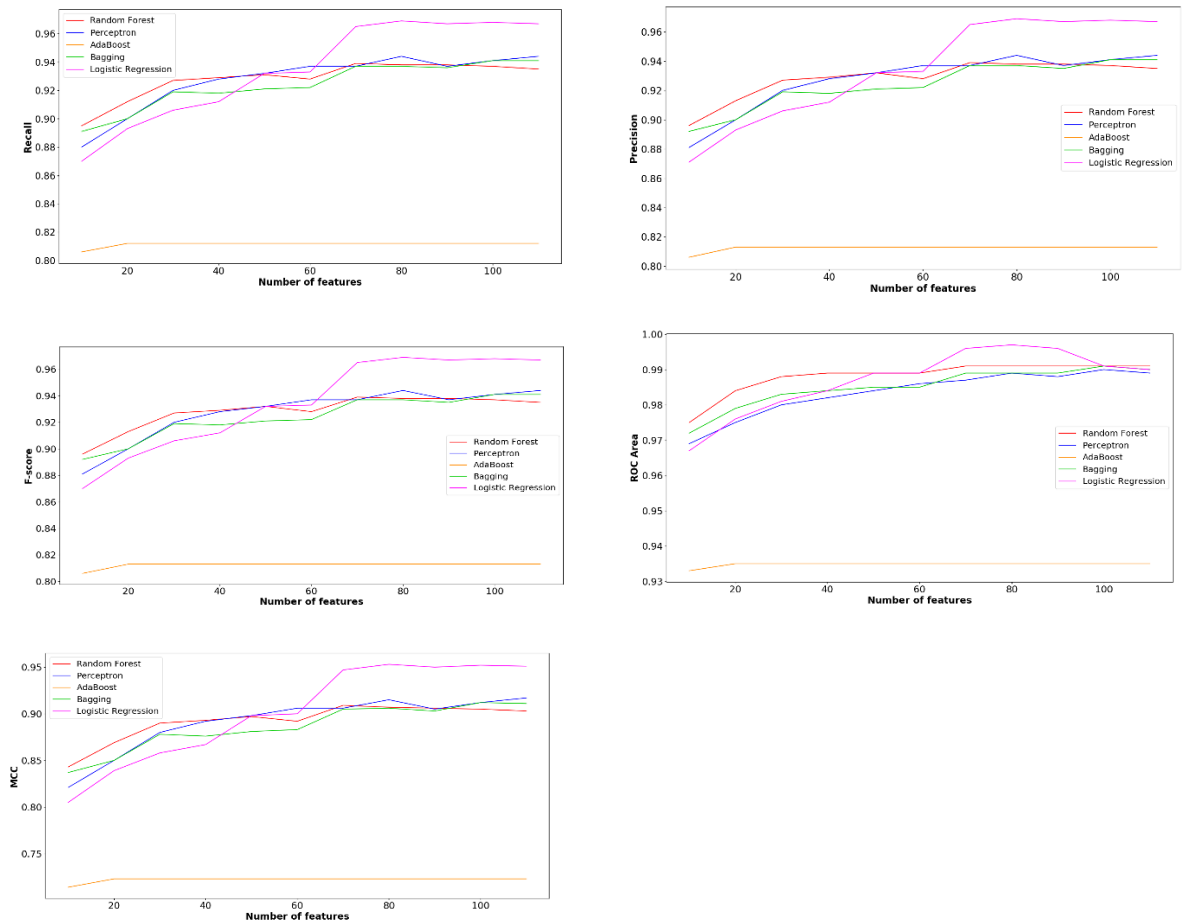

Figure 18. Effect of the number of selected features on text classification rates (reliefF).

A second experiment, which extends the previous one, aims at showing the correlations between three variables: the feature selection methods, the number of selected features, and the ML algorithms (figure 19). Here as well, on the x-axis, the features are ranked and grouped by 10 from highest to lowest according to the four adopted methods of feature selection. The F-score is adopted as it combines recall and precision. On the other hand, as shown in figure 15, the F-score results are usually between those of MCC and ROC, this makes them a good indicator of the actual performance.

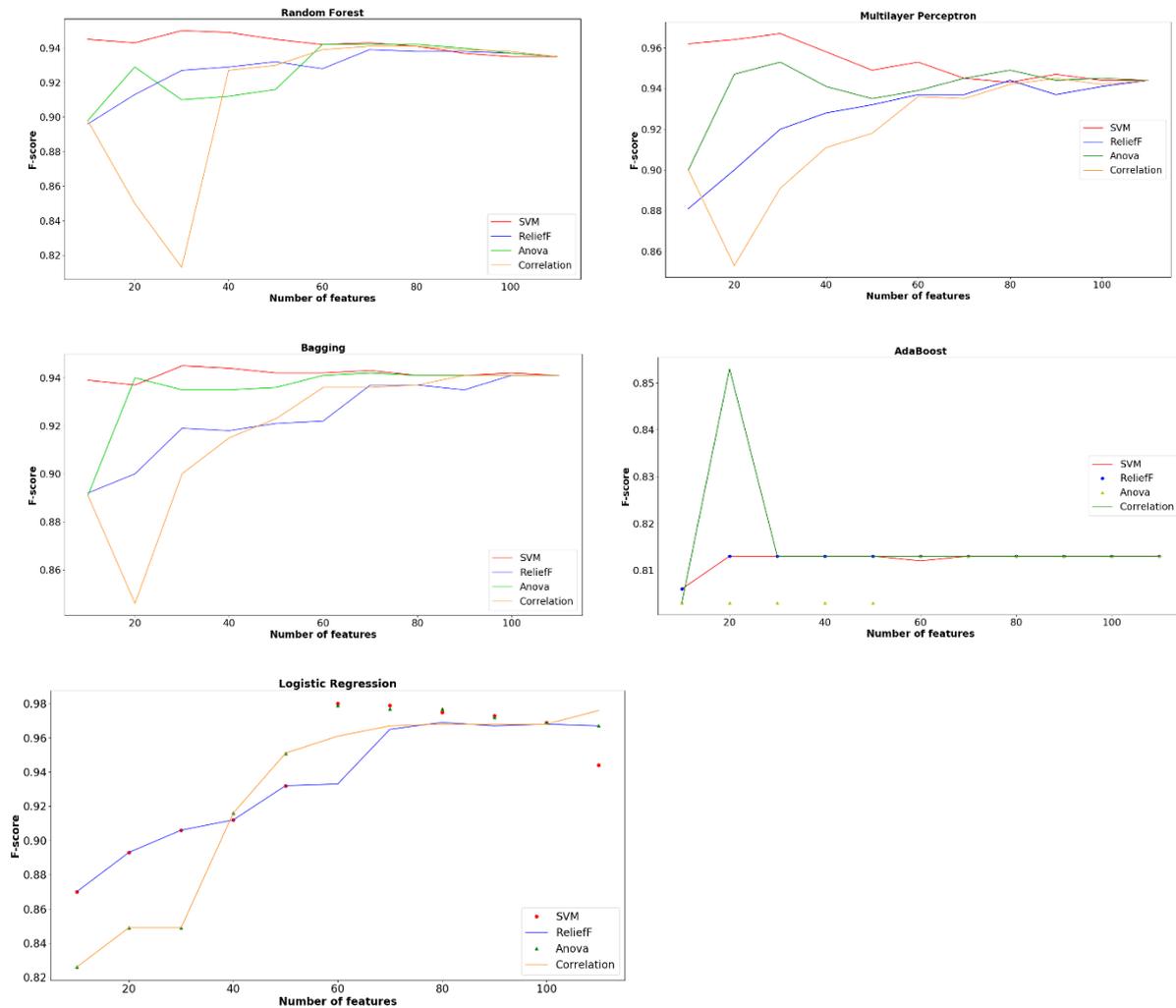

Figure 19. Comparison of the F-scores per number of features with the four adopted feature selection methods.

## 12.4 Feature selection by linguistic area

A second approach to feature selection consists of using linguistic areas. The goal of this experiment is to evaluate the importance of these groups of features in the classification process although these results should not be viewed as an exact comparison between the roles of the linguistic areas, as the covered features in this study are not exhaustive.

To measure the importance of the six linguistic areas that are covered (phonology, morphology, lexicon, syntax, discourse, and psychology), the percentages of the features selected from each of these areas are compared across the four adopted feature selection methods (figure 20). In this experiment, only the 50 features with the highest feature selection scores are considered.

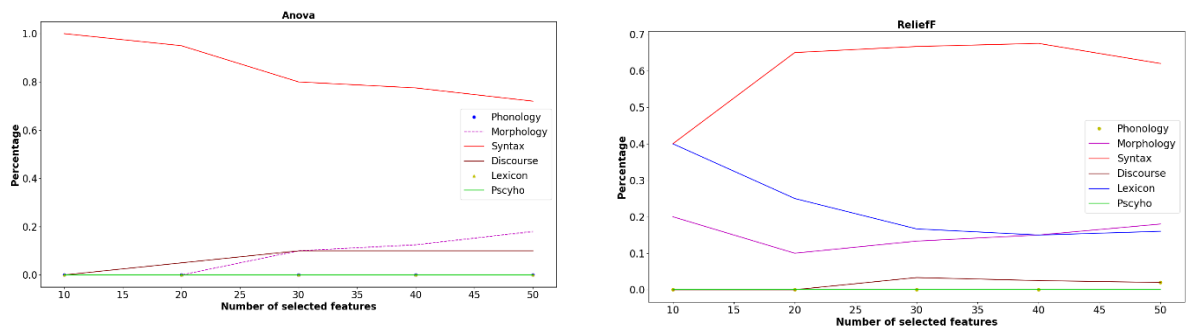

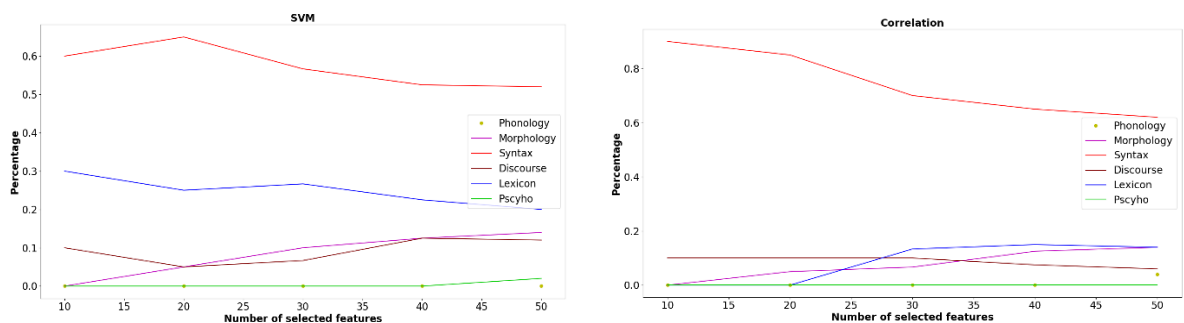

Figure 20. Percentage of selected features per linguistic area and per feature selection method among the 50 best-selected features.

To get a better idea about the role of every linguistic group of features, a comparison of the F-scores with each of these groups using the five adopted ML algorithms is depicted in figure 21.

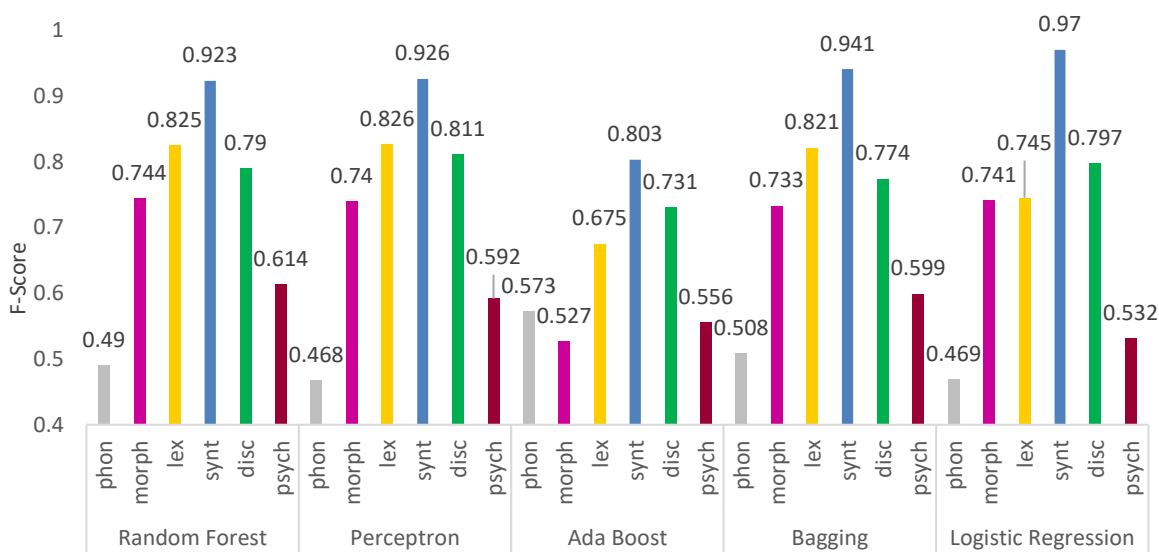

Figure 21. F-score results per linguistic area[21].

## 12.5 Readability Formulas

As seen in table 5, the readability formulas are a combination of different linguistic features, therefore combining them with the linguistic features would be a redundancy. However, it is interesting to test how a single readability formula can classify the text's linguistic complexity and how these formulas compare with the linguistic features presented in this paper. In figure 22, the F-scores of the classification of the text's complexity are depicted using the different readability formulas alone and combined, using the different ML algorithms.

---

[21] In this figure, *phon* stands for phonology, *morph* for morphology, *lex* for lexicon, *synt* for syntax, *disc* for discourse, and *psych* for psychology.

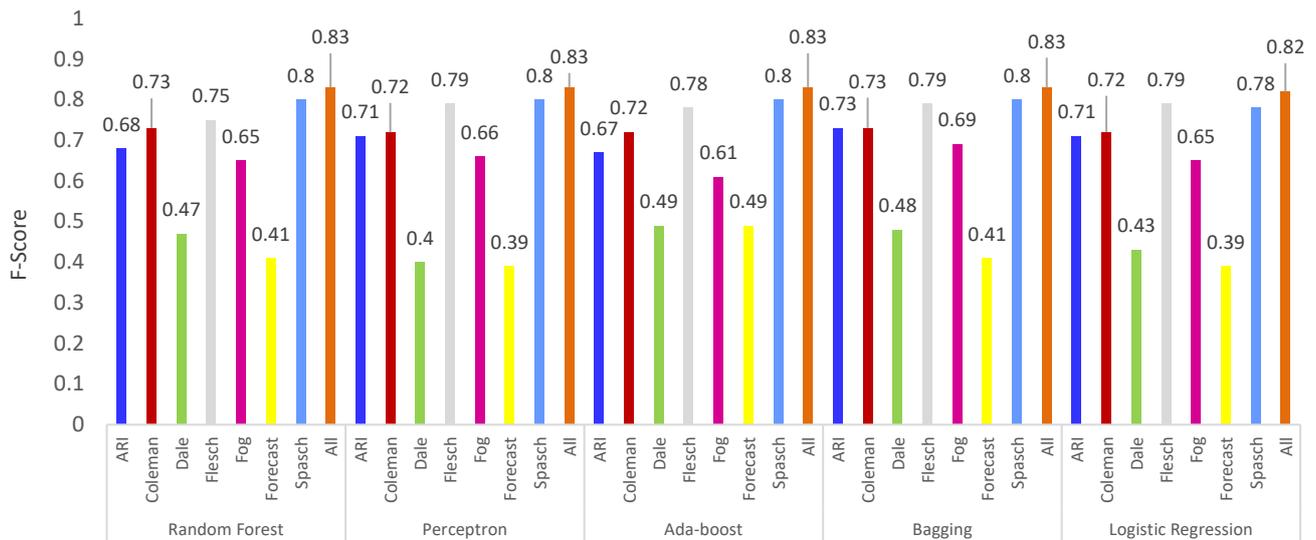

Figure 22. F-score of the different readability formulas alone and combined using the different ML algorithms[22].

## 12.6 Training Time Experiments

A related aspect of feature selection is the time necessary to build the model. The goal of this experiment is to show how the number of used features influences the training time with the five adopted ML algorithms. The reliefF feature selection method is adopted in this experiment. In figure 23, the times, in seconds, necessary to build the ML models are reported. These times are obtained using Weka 3.8.3 on a desktop with a 64-bits Windows 10, an i3 processor, and 8 GB RAM memory.

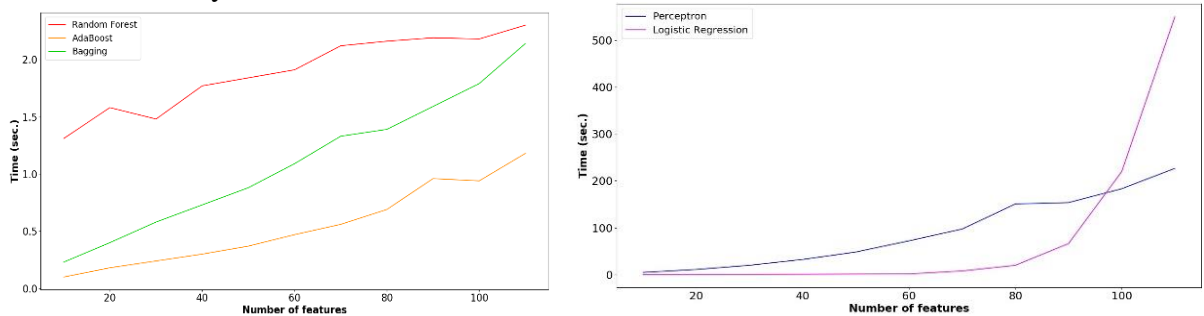

Figure 23. Training time comparison between the ML algorithms, time in seconds using reliefF.

One can look at the training time as a function of the interaction between the feature selection methods and the ML algorithms. In figure 24, the training times per number of selected features are depicted for the five adopted ML algorithms with the four used feature selection methods.

---

[22] Here is a list of shortcuts adopted in this figure: Coleman for Coleman-Liau, Dale for Dale-Challe, Fog for Gunning's Fog, and Flesch for Flesch-Kincaid, Spach for Spache.

Journal of Data Mining and Digital Humanities                                http://jdmdh.episciences.org
ISSN 2416-5999, an open-access journal

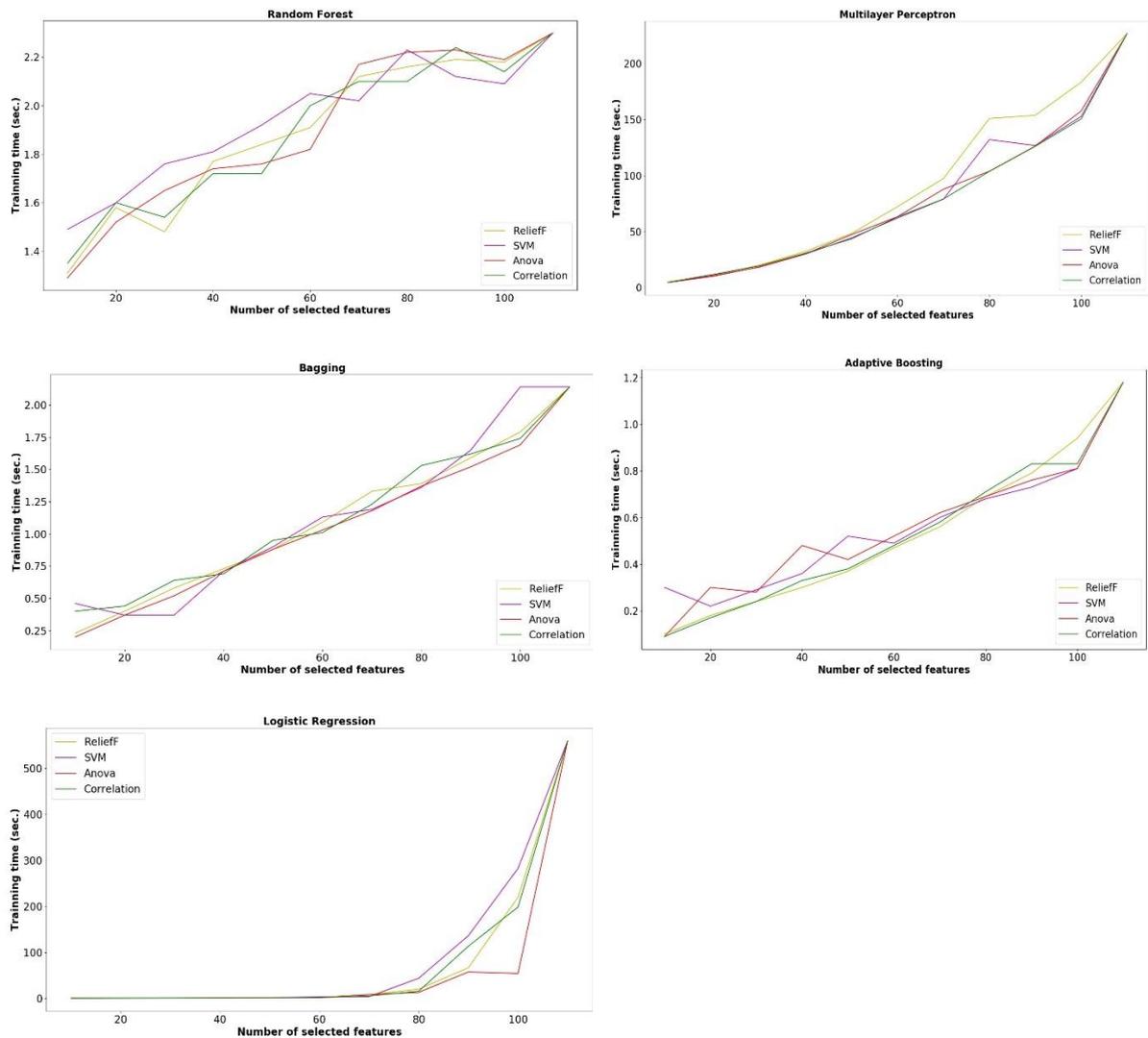

Figure 24. Training time comparison across the ML algorithms and the feature selection methods.

## XIII DISCUSSION

### 13.1 Overall Performance

As shown in figure 15, Logistic Regression provides the highest classification scores across the adopted measures, while Bagging, Multilayer Perceptron, and Random Forest have similar results. Finally, Adaptive Boosting provides the lowest scores across the adopted measures. The limited size of the training data can explain this low performance. In a different work about authorship attribution based on linguistic complexity, using the Enron corpus, a large dataset with about a half-million emails [Kurdi, 2019], the Adaptive Boosting provided very high scores outperforming Logistic Regression, Multilayer Perceptron, and Random Forest.

Besides, it is also important here to do an analysis of the confusion between the classes to understand the effect of the data on the classification rate. The confusion matrix with the two ML algorithms with the highest performance is presented in table 7.

|  | Actual | classified 1 | | classified 2 | | classified 3 | | Total |
|---|---|---|---|---|---|---|---|---|
| Multilayer | 1 | **1993** | 96.8% | 58 | 2.8% | 7 | .3% | 2058 |
| Perceptron | 2 | 53 | 2.5% | **1922** | 92.1% | 110 | 5.2% | 2085 |
|  | 3 | 6 | 0.2% | 109 | 5.3% | **1913** | 94.3% | 2028 |
|  | 1 | **2013** | 97.8% | 42 | 2% | 3 | 0.1% | 2058 |
| Logistic | 2 | 49 | 2.3% | **1986** | 95.2% | 50 | 2.3% | 2085 |
| Regression | 3 | 8 | 0.39% | 49 | 2.4% | **1971** | 97.1% | 2028 |

Table 7. Confusion matrix of the Multilayer Perceptron and the Logistic Regression with all the features, in every column the numbers and the percentages of the cases are provided.

As we can see in table 7, with both algorithms, class 1 is mostly confused with class 2 but rarely with class 3. Class 2, being in the middle, is almost equally confused with class 1 and class 3 with Logistic Regression. With the Multilayer Perceptron, class 2 is confused about twice as much with class 1. This shows that with both ML algorithms, the overall behavior is coherent and fits the data well.

Finally, it is hard to compare this performance to other works about text classification by linguistic complexity given the differences in the used data sets. However, the results reported in this paper are superior to those reported in [Davoodi, 2016], [Feng et al., 2010], [Vajjala, 2013], [Xia et al., 2016], [Kurdi, 2017a], and [Balyan et al., 2018].

### 13.2 Scalability Experiments

Adding an extra level of text complexity (level four) did not cause the system performance to decrease, except with the Adaptive Boosting (figure 16). In this second experiment, a significant decrease in performance is observed: the F-score went down to about 65 with Random Forest and bagging (figure 17). However, a careful analysis of the results through the confusion matrix gives a more positive description of the situation (table 8). Only one text of level four is taken as a text of level three. This good performance occurs even though the training and testing are done on two different types of corpora for level four (journalistic texts vs. college student essays). Besides, all the texts of level one were correctly classified. The texts of level two have the highest confusion rate. This confusion is expected, given the proximity of texts of level two with the two other classes from both sides and given the potential misalignment of the levels between the ESLTL and CCT corpora.

| Actual | classified 1 | | classified 2 | | classified 3 | | classified 4 | |
|---|---|---|---|---|---|---|---|---|
| 1 | **50** | 100% | 0 | 0% | 0 | 0% | 0 | 0% |
| 2 | 11 | 0.22% | **20** | 40% | 8 | 16% | 11 | 22% |
| 3 | 0 | 0% | 10 | 20% | **17** | 34% | 23 | 46% |
| 4 | 0 | 0% | 0 | 0% | 1 | 2% | **49** | 98% |

Table 8. Confusion matrix of the Random Forest with all the features (second scalability experiment).

Overall, one can conclude that the system has a coherent performance, as the confusion is limited between texts of contiguous levels.

### 13.3 Automatic Feature Selection Experiments

As seen in figure 18, using the reliefF feature selection method and adding more features can lead to a positive, negative, or neutral effect on the classification performance. Note that the effect is not the same across the five adopted performance measures. For example, adding the last eleven features causes the ROC area score of the Logistic Regression algorithm to go down, while the values of the other measures remain the same with this algorithm. Besides, recall and precision have similar curves. Furthermore, Adaptive Boosting is the least ML to be affected by the number of features: after a slight performance improvement at 20 features, the performance remains constant. This is because this ML algorithm builds weighted classifiers.

This process acts as an internal feature selection functionality. Adaptive Boosting was even used as a method for feature selection [Wang, 2012], [Redpat and Lebart, 2005]. Finally, with reliefF, 70 is the number of features beyond which the performance of the different ML algorithms stagnates, decreases, or improves marginally across all the performance measures. The performance of the different ML algorithms is compared with the four feature selection methods in figure 19. This figure shows that the feature selection methods and the ML algorithms interact differently. For example, with the correlation method, the features 31-40 lead to performance improvement with Logistic Regression while the same features lead to a performance drop with bagging. Besides, with the Multilayer perceptron, bagging, and random forest, the SVM feature selection method gives high F-scores with 10 features only (between 0.94-0.96). On the other hand, with the same three algorithms, the ANOVA effect size gives its best performance with the best 20 features. Adaptive Boosting displays an exceptional behavior, as its performance does not change after 30 features regardless of the feature selection method. This can be explained by the fact that this algorithm does its own feature selection while building the optimal classifier as explained before. Finally, figure 19 confirms that the performance of ANOVA's based omega squared is a viable approach to feature selection, as it provides the highest performance with the Bagging ML, with 20 features.

**13.4 Feature Selection by Linguistic Area**

As seen in figure 20, there is a disagreement between the different feature selection methods about the selected percentages of the linguistic areas. For example, ANOVA and Correlation rely mostly on syntax with some percentages from morphology and discourse, while SVM and reliefF use more diversified features. This is probably because ANOVA does not consider the correlations between the features. Besides, there are some general tendencies across the feature selection methods. For example, all the methods rely heavily on syntax in their 50 best features, while they agree on not using psychology and phonology features.

As seen in figure 21, the syntactic features give the highest F-scores across the ML algorithms, while the phonology features give the lowest F-scores with all the ML, except with Adaptive Boosting. The big difference in terms feature's numbers between the two disciplines (three phonology features and fifty-two syntactic features) is the reflection of their role. For example, phonology is mainly concerned with spoken language. It is interesting to note that syntax alone gives a similar F-score rate than all the features combined. On the other hand, the role of the twelve lexicon features varies depending on the ML algorithm as it plays a strong role with Random Forest, Multilayer Perceptron, and Bagging (F-score > 0.80) while it plays a moderate role with Adaptive Boosting (F-score = 0.67). This role is observed despite the high redundancy of some lexical features like CTTR, GTTR, and TTR. The twelve discourse features have a homogeneous performance across the ML algorithms, which is around 0.73-0.81. With Adaptive Boosting and Logistic Regression, the discourse has the second-highest F-score after syntax. While, with the Multilayer Perceptron, discursive features have an F-score close to the lexicon that has the second-highest score. Morphology, with its twenty features, plays a varied role in the classification across the algorithms. The F-score is only 0.52 with Adaptive Boosting, while it is around 0.74 with the other ML algorithms. Finally, the 12 psychology features have F-scores between 0.53 and 0.61. These low scores are justified because these features measure aspects that are more dependent on the content than on the linguistic form, such as the meaningfulness, the imageability, and the concreteness.

**13.5 Readability Formulas**

Given the high redundancy of the readability formulas as shown in table 6, using all the readability formulas does not make a big improvement compared to the totality of the features. For example, Spache, the readability formula with the highest performance, gives an F-score of about 0.80, while all the readability formulas combined give a score that is about 0.83. Although Dale-Challe, Gunning's Fog, and Spache cover similar linguistic aspects: sentence length and frequency of words, the F-score of Spache is significantly higher than Gunning's Fog. This difference shows that the weighting of the variables in the readability formulas plays an important role. Besides, Flesch-Kincaid, which uses a combination of sentence complexity and word complexity, has the second-highest performance.

In terms of ML algorithms, there is no ML algorithm that gives systematically the highest or lowest F-scores across the different readability formulas. For example, Logistic Regression gives one of the highest scores with Flesch-Kincaid and one of the lowest with FORCAST. For all the readability features combined, Logistic Regression gives a slightly lower F-Score 0.82 than the other ML algorithms, which is around 0.83. This result shows, from a practical point of view, that the ML algorithms are not only sensitive to the number of features and the data set size, but also to the nature of the features. Finally, the results reported in figures 22 and 15 show that the linguistic features outperform significantly a single or combined linguistic readability formulas. This result also confirms the experiments conducted on a smaller number of readability formulas with linguistic features by [Crossley et al., 2011].

### 13.6 Training time experiments

The comparison of the training times, done in figure 23, shows that across the ML algorithms adding new features leads to an increase or a stagnation in training time. The overall performance of the chosen ML algorithms shows that the Multilayer Perceptron and the Logistic Regression, which are simple ML algorithms, need a much higher time for training than the three adopted ensemble ML algorithms. Neural networks are known for their slow training time. This is confirmed with the Multilayer Perceptron whose training time is the highest within the range 10-80 features. The training time amount of increase varies from ML algorithm to another. For example, for the Logistic Regression algorithm, the training time increases more significantly starting from 80 features.

Among the ensemble algorithms, Random Forest requires systematically the largest running time; Adaptive Boosting requires systematically the least amount of time while bagging is in the middle.

As seen in figure 24, the behavior of the ML algorithms in terms of training regarding the number of selected features is not the same. For example, Logistic Regression shows small increases in the training times between the range 10-70 features, and then a big increase starts to occur, with minor differences between the different feature selection methods. Logistic Regression is known for its slowness with learning large numbers of variables. The homogeneous behavior of the Multilayer Perceptron with the increase of the number of features can be explained by the fact that the number of neurons in the input layer is equal to the number of features. In other words, the architecture of the NN is adjusted to fit the data. Additionally, Bagging, Adaptive Boosting, and Random Forest, which are ensemble algorithms, have different training times' patterns. Random Forest curves have more fluctuations than Bagging and Adaptive Boosting. The explanation of this difference may lie in the internal mechanisms of these three algorithms for building their ensemble of weak classifiers. With Bagging and Adaptive Boosting, the classifiers are built with sampled data sets of the same size. However, with Random Forest the size of the sub-data set is random.

### XIV CONCLUSION AND PERSPECTIVES

This paper is about the exploration of 118 features of different linguistic areas in addition to different readability formulas within the context of ESL text classification by linguistic complexity. Although most of the explored features are already proposed in the ESL and text-mining literature, some of these features, like the Continuous Lexical Sophistication CLS, verb tenses, the usage of POS tags ngram profiles and implementing word embedding based coherence features were proposed for the first time in this paper. Besides the focus on the classification side, this paper also offers an evaluation of the individual contribution of every one of these features through their effect sizes, a discrete translation of ANOVA's F-score. The experiments were conducted using a corpus of 6171 texts from six professional ESL websites. The overall performance is good, with an F-score 0.97, obtained with the Logistic Regression ML algorithm and all the features. The infrequent errors that occurred are between texts of contiguous levels (e.g. texts classified as level two instead of level three), which confirms the coherent behavior of the classifiers.

A scalability evaluation was conducted to test if such a classifier would be used within real applications where it can be, for example, plugged into a web scraping module. In this evaluation, the texts in the test set are not only unused in the training but also of different types (ESL texts vs. children texts). Although the overall performance of the classifier decreased significantly, the confusion matrix shows that most of the classification errors are between the classes two and three (the middle classes) and that the system has a robust performance in categorizing texts of class one and four. This behavior can be explained by the difference in classification criteria between the two corpora. Hence, the work presented here gives a good foundation for recommender systems that propose texts of the right linguistic complexity to ESL learners.

This paper also offered some insights about four different feature selection approaches and showed that the proposed ANOVA's based omega squared gave good results, especially for the 20 best features. Furthermore, this work showed that, in addition to syntax, other linguistic areas like discourse, lexicon, and morphology could play a central role in text complexity classification. Finally, it showed that the readability formulas that combine measures of sentence complexity and word complexity could achieve a decent F-score going up to 0.8 with Spache. Nevertheless, this score is significantly inferior to the one achieved by the linguistic features.

The training time experiments confirmed that Multilayer Perceptron is among the slowest learners. It also showed that logistic regression starts to be even slower than Multilayer Perceptron, when given data with ninety features or more.

In terms of perspectives, this work can be improved by adding more features like using Elmo [Peters et al., 2018] and Bert [Devlin et al., 2019] for word embedding to measure discourse coherence. Another feature worth exploring is the lexical variation over the n-grams that was proposed by [Ramirez de la Rosa et al., 2013]. The wide coverage of the features explored here can apply to a wide range of areas like authorship attribution, dementia detection, and language acquisition delay detection. Also, in addition to the discrete recommendations by text level covered in this paper, the usage of the features studied here have been explored within a more continuous adaptive recommender system [Kurdi, 2018]. In such a recommender, the system proposes the best-suited text based on users' profiles that are built with the features extracted from a read text. The recommended texts are the closest unread texts in an n-dimensional space, where *n* is the number of extracted features.

**Acknowledgement**

I would like to offer my special thanks to professor Bin Zou (Xi'an Jiaotong-Liverpool University) and to the anonymous reviewers for their constructive feedback about this paper.